\def\mycmd{2}   
\if\mycmd1  
\documentclass[12pt, draftclsnofoot, onecolumn]{IEEEtran}
\else       
\documentclass[a4paper, 10pt,journal]{IEEEtran}
\IEEEoverridecommandlockouts
\fi         

\ifCLASSINFOpdf
\else
\fi

\usepackage{graphicx}

\usepackage{amsmath}
\usepackage{amssymb}
\usepackage{authblk}
\usepackage{comment}
\usepackage{cite}
\usepackage{algorithm}
\usepackage{tabularx}
\usepackage{multirow}
\usepackage[noend]{algpseudocode}
\usepackage[margin=0.75in]{geometry}
\usepackage{hyperref}
\hypersetup{
    colorlinks=true,
    linkcolor=black,
    filecolor=magenta,      
    urlcolor=blue,
    citecolor =black,
}

\algnewcommand{\Initialization}[1]{%
  \State \textbf{Initialization:} \hspace{0.13in}
}
\makeatletter
\newcommand{\multiline}[1]{%
  \begin{tabularx}{\dimexpr\linewidth-\ALG@thistlm}[t]{@{}X@{}}
    #1
  \end{tabularx}
}
\makeatother

\newcommand{\rom}[1]{\uppercase\expandafter{\romannumeral #1\relax}}

\newcommand*{\argmin}{\operatornamewithlimits{argmin}\limits}

\begin{document}


\title{Privacy-Protection Drone Patrol System based on Face Anonymization}

\author{Harim Lee$^{1}$, Myeung Un Kim$^{2}$, Yeongjun Kim$^{1}$, Hyeonsu Lyu$^{1}$, and Hyun Jong Yang$^{1}$
\thanks{$^{1}$Ulsan National Institute of Science and Technology, Ulsan, South Korea (e-mail: \{hrlee,hjyang\}@unist.ac.kr)}%
\thanks{$^{2}$Korea Institute of Science and Technology, Seoul, South Korea}%
\thanks{Hyun Jong Yang is the corresponding author.}
}

\maketitle
\begin{abstract}
The robot market has been growing significantly and is expected to become 1.5 times larger in 2024 than what it was in 2019. Robots have attracted attention of security companies thanks to their mobility. These days, for security robots, unmanned aerial vehicles (UAVs) have quickly emerged by highlighting their advantage: they can even go to any hazardous place that humans cannot access. For UAVs, Drone has been a representative model and has several merits to consist of various sensors such as high-resolution cameras. Therefore, Drone is the most suitable as a mobile surveillance robot. These attractive advantages such as high-resolution cameras and mobility can be a double-edged sword, i.e., privacy infringement. Surveillance drones take videos with high-resolution to fulfill their role, however, those contain a lot of privacy sensitive information. The indiscriminate shooting is a critical issue for those who are very reluctant to be exposed. To tackle the privacy infringement, this work proposes face-anonymizing drone patrol system. In this system, one person's face in a video is transformed into a different face with facial components maintained. To construct our privacy-preserving system, we have adopted the latest generative adversarial networks frameworks and have some modifications on losses of those frameworks. Our face-anonymzing approach is evaluated with various public face-image and video dataset. Moreover, our system is evaluated with a customized drone consisting of a high-resolution camera, a companion computer, and a drone control computer. Finally, we confirm that our system can protect privacy sensitive information with our face-anonymzing algorithm while preserving the performance of robot perception, i.e., simultaneous localization and mapping.


\end{abstract}

\IEEEpeerreviewmaketitle

\section{Introduction}
\label{sec:intro}
The security robot market is expected to grow from USD 2.106 billion in 2019 to USD 3.33 billion by 2024 \cite{Security_Robot_Makret}.
As mobile surveillance devices, security robots are slowly being deployed in several common sights such as malls, offices, and public spaces.
Since mobile surveillance devices consist of artificial intelligence, cameras, and storage, they can more reliably collect data than human, suggesting the security robots can effectively replace human security guards.

For security robots, there are several robot types: unmanned ground vehicles and unmanned aerial vehicles.
Especially, the unmanned aerial vehicles attract the attention as security patrol robots since those can move anywhere more freely than the unmanned ground vehicles.

Recently, Drone has been emerging in unmanned aerial vehicles.
The drone market is expected to reach USD 129.23 billion by 2025 \cite{Drone_Market}.
Thanks to the capability to go any place that humans and ground vehicles cannot reach, Drone rapidly has been taking the place of the conventional surveillance methods such as fixed CCTV and human guards \cite{Drone_Market_2}.
In addition, since Drone can consist of high-computing onboard computer, cameras, and several sensors, those have considerable potential to increase surveillance by adopting high-quality image processing.

However, security patrol robots incur several issues such as privacy infringement as well as malfunction. 
Since Drone can go anywhere with high-resolution cameras, it potentially has a damaging threat to privacy.
The representative UAV takes videos in public spaces with a high-resolution camera to fulfill their role, and then the indiscriminate shooting is a crucial problem for people who are disinclined to be exposed and highly regard their privacy.
A drone indeed can be used to infringe a person's privacy with a camera spying on a person \cite{Privacy_Fear}.
In 2015, a Kentucky man shot down a drone hovering over his property \cite{shoot_down_drone, shoot_down_drone_2}.
He argued that the drone was spying on his 16-year-old daughter who was sunbathing in the garden.
In addition, an article reported that a Knightscope security robot was suspended from its job of patrolling a San Francisco animal shelter after some residents felt the robot was taking unnecessary photos of them \cite{Privacy_Infringement_1}.

Due to robot cameras, human privacy infringement issues could become a critical but pervasive social problem \cite{Privacy_concern}.
Hence, a clever solution should be developed to preserve the person's privacy, but require no sacrifice of robot perception performance.
Several works have proposed some methods to protect the privacy invasion \cite{PlaceAvoider, Jason, MU_Low_Anonymizer}.
Those schemes detect privacy-sensitive information from images, and then remove or anonymize it via machine learning techniques.

This work proposes a privacy-preserving drone patrol system with face-anonymizing networks.
In our face anonymizing framework, the key idea is to modify one face in a video frame to look like it is not his/her face.
In our system, the face-anonymizing framework is implemented on a companion computer which is loaded on a drone.
Hence, video frames recorded by a drone's camera are immediately processed on the companion computer by our face-anonymizing networks, and then the results will be transmitted via a wireless technology. 
Specifically, the contribution of this work is summarized as follows.
\begin{itemize}
    \item We propose our face-anonymizing approach, where a face image is transformed to an intermediate image by removing the privacy-sensitive information, and then the intermediate image is converted to a photorealistic face image.
    \item To realize our approach, this work presents a training architecture to combine two latest generative adversarial networks (GANs).
    Then, we explain challenges to train deep learning networks for our system purpose.
    Finally, it is introduced how to overcome the challenges.
    Note that our proposed modifications' approach is not limited to the adopted training frameworks, but can be utilized in any other GAN frameworks for anonymizing purpose.
    \item Via various face image and video dataset, we conduct extensive evaluation to verify our face-anonymizing framework. The results confirm that our proposed scheme anonymizes faces in various images and videos well enough.
    \item We build a drone consisting of a zed camera and a companion computer to demonstrate our drone patrol system. Our face-anonymizing framework is loaded on the companion computer. 
    We utilize robot operating system (ROS) framework to connect each component in our system.
    Via a real video recorded by a drone's camera, we present how our system will actually work in reality.
\end{itemize}


\subsection{Related Work}
\subsubsection{Removal of Privacy Sensitive Information}
In robot cameras, the privacy infringement has attracted attention to develop a method removing the privacy-sensitive information in images \cite{PlaceAvoider, Jason, MU_Low_Anonymizer}.

In \cite{PlaceAvoider}, the authors introduced scene recognition from a image.
The scheme determines if a person is in a privacy-sensitive location.
If a image is taken in a privacy-sensitive place, the proposal allows a camera device to be automatically turned off. 
However, faces are still exposed in privacy-insensitive places, and thus this scheme is not suitable for patrol drone visions.

Jason et al. \cite{Jason} developed the privacy preserving action detection via a face modifier by using Generative Adversarial Networks (GANs).
They proposed pixel-level modifications to change each person's face with minimal effect on action recognition performance. 
The proposed generator modifies each pixel in a original face to remove features of the face.
To train the generator, the authors design their discriminator based on a face identification network that recognizes who s/he is.
The generator learns the way to make the discriminator believe that a generated face is different from the original image.
However, the generator tends to replace a input face to another face in the training dataset.
In other words, the problem is that the generator doesn't learn how to change a face not in the training dataset.
Moreover, the generator observes pixels of a face for modification, which means that a modified face is generated based on the original face.
This approach could still leave some information of a original face in a modified face.

The work \cite{MU_Low_Anonymizer} proposed a dynamic resolution face detection architecture to blur faces. 
The framework detects faces from extreme low resolution images via the proposed deep learning-based algorithm.
Except for the detected faces, other privacy-insensitive pixels are enhanced to high resolution.
Hence, in result images, only faces are blurred, which protects privacy-sensitive parts while preserving the performance of robot perception.
However, in the case that a face are big in a frame, an intimate person can recognize who the person is in the frame even if the face is blurred.
In addition, it would be possible not to detect a face in a low-resolution image, and then this scheme couldn't protect a person's privacy.

\subsubsection{Generative Adversarial Networks}
Generative Adversarial Networks (GANs) have had impressive success in generating realistic images \cite{GAN_Goodfellow}.
The goal of this learning framework is to train a neural network to model a image distribution in an unsupervised manner.
The trained network can generate a fake image indistinguishable from a real image.
This training approach have been adopted to image-to-image translation \cite{Huang_18, Isola_17, Karacan_16, Karacan_18, Liu_17, Zhao_19, Zhu_17, Zhu_17_2, Park_19}.
Those works learn a mapping from input to output images, meaning that a input image is translated to a image in a different image distribution.
To construct our training architecture for obtaining deep-learning networks for our purpose, we have adopted the latest two works \cite{Park_19, Zhu_17}.
By using the training framework in \cite{Zhu_17}, we make a generator to translate a photorealistic image to a segmentation mask, and the work \cite{Park_19} is used to train another generator that converts the resultant segmentation mask to a photorelistic image.

\subsubsection{SLAM} To verify that our anonymization method has no effect on vision-based robot perception, we utilize simultaneous localization and mapping (SLAM) techniques.
By using SLAM, in an unknown environment, a robot constructs a map around itself and localizes itself in the resultant map.
Hence, in a video frame, manipulation of some pixels could affect the performance of SLAM since a map is drawn by extracting feature points of lines, edges, and corners of objects in images.
In this work, ORB-SLAM2 \cite{ORB-SLAM2} is implemented on our system, which is one of the most popular algorithms of the vision-based SLAM.


\section{Privacy-Protection Drone Patrol System}
\label{sec:main}
\begin{figure*}[!t]
\centering{}\includegraphics[width=0.7\paperwidth]{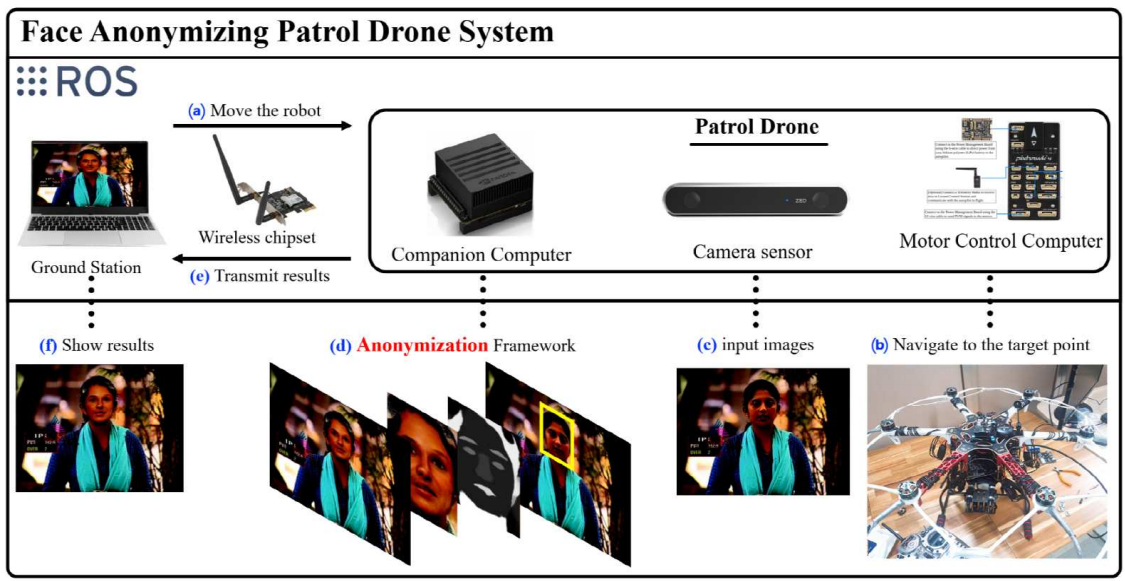}\centering\protect\caption{Composition of the developed patrol robot system with privacy preserving face detection.}
\label{Fig:framework}
\vspace{-0.1in}
\end{figure*}

This section introduces our face anonymizing drone patrol system in Fig.~\ref{Fig:framework}, which consists of a ground station and a video-recording drone.

\textbf{Ground Station}: this component has two tasks: (1) command and (2) viewer.
The ground station is connected to the video-recording drone via Wi-Fi, one of the wireless technologies.
This part is running a command program, and then controls location of the drone.
Via the program, we can order the drone to move toward a specific point.
In addition, through wireless communication, this ground station receives resultant video frames from the drone, and then presents those frames via our viewer.

\textbf{Video-Recording Drone}: 
The drone consists of a high-resolution camera, a companion computer, and a motor control computer, where the camera and the motor control computer are connected to the companion computer.
The companion computer fetches video frames from the camera, and then anonymizes faces in the received frames by executing our face-anonymizing networks.
The ORB-SLAM2 is additionally processed in the resultant videos.
The final results are transmitted to the ground station.
In addition, the companion computer receives a moving command, and passes on it to the motor control computer.

The key features of our system are summarized as follows.
\begin{itemize}
    \item In our face-anonymizing system, deep-learning networks anonymize all detected faces never to be distinguishable for protecting the person's privacy.
    To obtain networks for our system purpose, our training architecture consists of two state-of-the-art GANs, called CycleGAN and GauGAN.
    Moreover, we modify the generator's and discriminator's loss of both GANs.
    \item In our system, the anonymization process is performed on a companion computer in a drone, which fetches video frames directly from a high-resolution camera.
    Hence, the transmitted video frames have no privacy-sensitive information, and thus a person's privacy can be completely protected.
    \item By implementing ORB-SLAM2 in our system, we verify that our anonymization scheme requires no sacrifice on the performance of vision-based SLAM.
    By anonymizing faces, our system effectively protects the person's privacy while preserving performance of the robot perception.
\end{itemize}

\section{Approach, Modifications, and Algorithm}
\label{sec:approach_modifications}
In this section, we introduce our approach for anonymizing faces via neural-type networks and a training architecture to train those networks. 
For our training architecture, we present how to combine two up-to-date GANs, and then We describe our modifications on losses for our purpose.
In addition, we explain the training procedure and how to make our training dataset.
Finally, we present and explain our face-anonymizing algorithm.

\subsection{Face Anonymizing Approach}
\begin{figure*}[t]
\centering
\includegraphics[width=0.7\paperwidth]{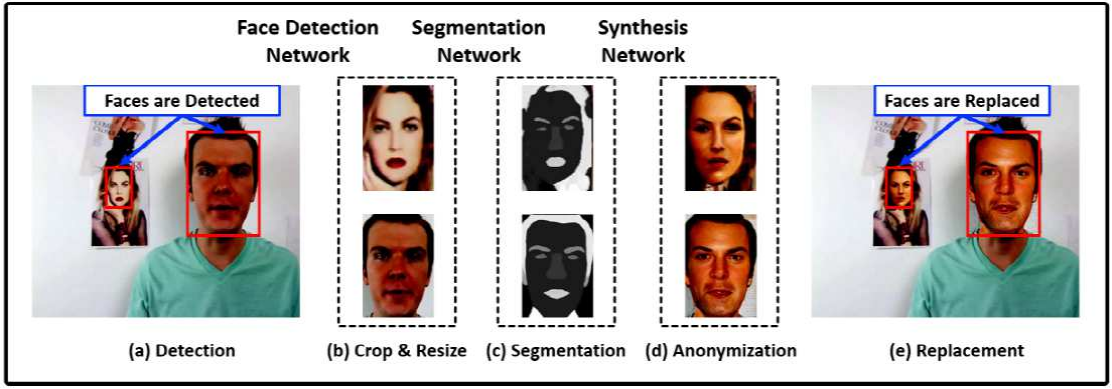}
\caption{Our Approach to Anonymize Faces in a Video Frame}
\label{Fig:Anonymization_Approach}
\end{figure*}
Fig.~\ref{Fig:Anonymization_Approach} illustrates our anonymization approach for our system.
To anonymize faces, a companion computer has three different networks: Face detection network, Segmentation network, and Synthesis network.
The face detection network operates to detect faces whenever a video frame is fetched to the companion computer.
Detected faces are cropped and resized to meet the required input size of the segmentation network.
Via the segmentation network, face's images are translated to semantic images.
Note that the semantic images have no privacy information but the outline of detected faces is still maintained in the resultant images.
The synthesis network generates photorelistic images based on the semantic images.
Finally, the phtorelistic images replace the original faces.
Since semantic images still have the outline of each facial component, phtorelistic images could retain facial expressions.

\subsection{Training Architecture for Segmentation and Synthesis Networks}
\if\mycmd1  
\begin{figure*}[t]
\centering
\includegraphics[width=0.8\paperwidth]{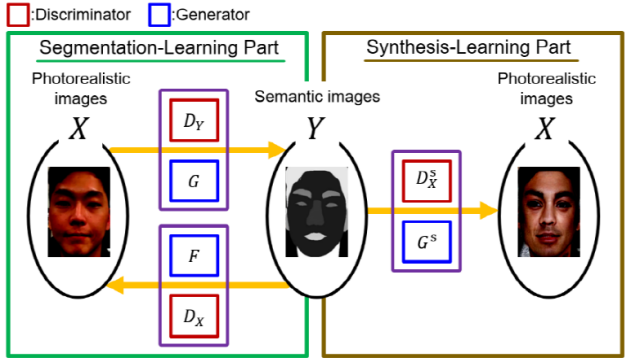}
\caption{Training Architecture for Segmentation and Synthesis Networks}
\label{Fig:Training_Architecture}
\end{figure*}
\else
\begin{figure}[t]
\centering
\includegraphics[width=0.99\columnwidth]{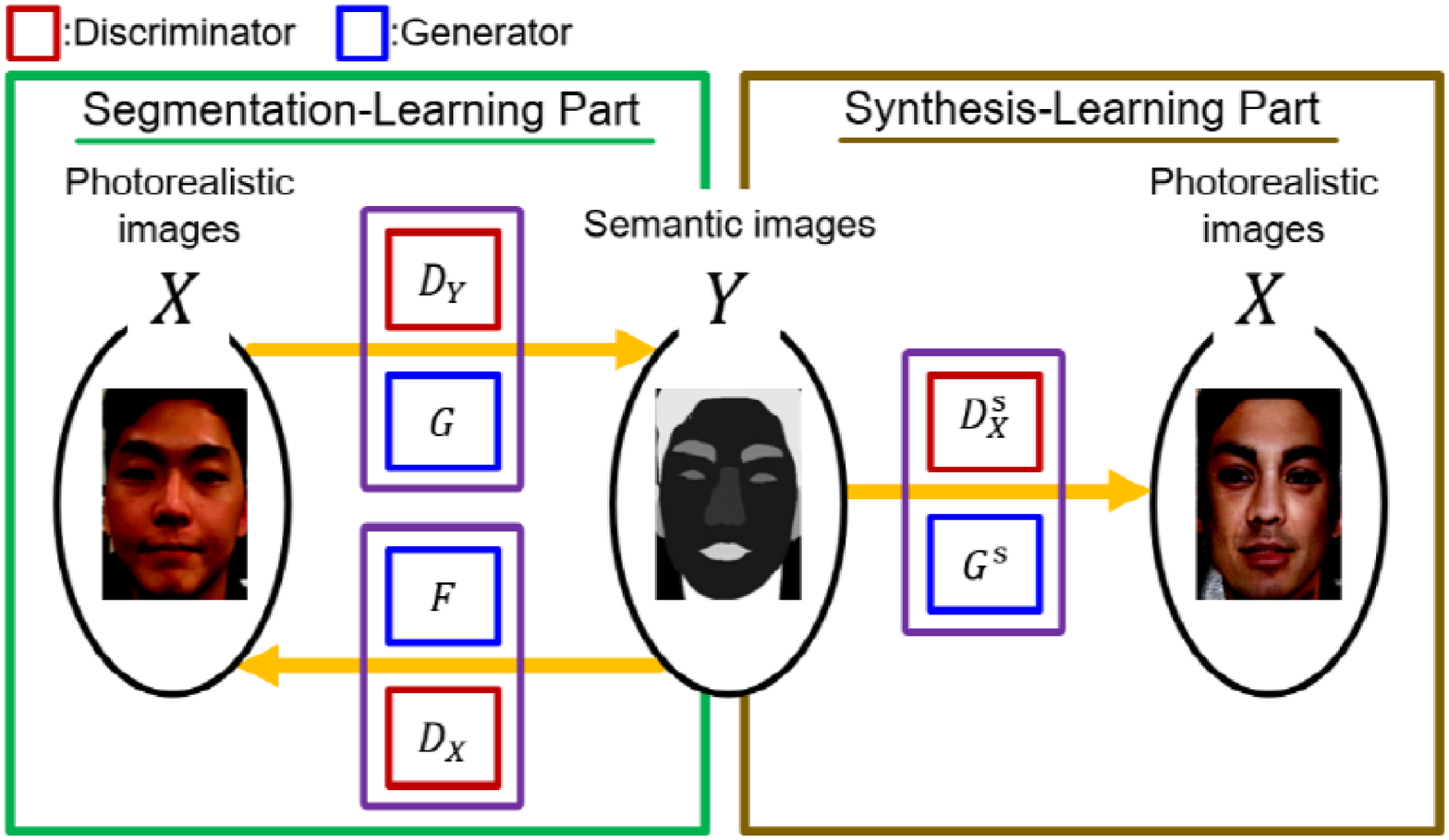}
\caption{Training Architecture for Segmentation and Synthesis Networks}
\label{Fig:Training_Architecture}
\end{figure}
\fi
Fig.~\ref{Fig:Training_Architecture} presents our training architecture for segmentation and synthesis networks.
To train those networks, we combine CycleGAN and GauGAN each of which is one of the spotlight image-to-image translation frameworks using GAN.
The segmentation network is trained by CycleGAN's framework, where Generator $G$ called segmentation generator is to generate a semantic images from a photorealistic image.
The synthesis network is obtained from GauGAN's framework.
Generator $G^\text{s}$ will make a photorealistic image from the output of segmentation generator $G$.

To obtain well-trained networks suitable for anonymization, we make modifications to generators' and discriminators' losses of both translation frameworks.

\subsubsection{Training Architecture Model}
To explain our modifications, we formulate our training architecture as follows.

For training samples, $X$ and $Y$ denote a photorelistic domain and a semantic domain, respectively.
For each domain, training samples are represented by $\{x_i\}_{i=1}^{N}$ and $\{y_j\}_{j=1}^{N}$.\footnote{In this work, $x_i$ is paired with a corresponding $y_j$, and thus both domains have the same number of samples.}
Samples of each domain are followed by a data distribution, which represents $x \sim p_\text{data}(x)$ and $y \sim p_\text{data}(y)$, respectively.
In this architecture, we have three generators, $G$, $F$, and $G^\text{s}$.
Each generator is a mapping function: $G: X \rightarrow Y$, $F: Y \rightarrow X$, and $G^\text{s}: Y \rightarrow X$.
For adversarial networks of those generators, there are discriminators $D_X$, $D_Y$, and $D_X^\text{s}$ each of which distinguishes if a input is from a data distribution or is generated by a generator.
$D_Y$, $D_X$, and $D_X^\text{s}$ examine output of $G$, $F$, and $G^\text{s}$, respectively.

\subsubsection{Well-Known Basic Definitions}
In this subsection, we introduce well-known definitions of losses \cite{Zhu_17, Park_19}.
Based on the losses, we will describe our modifications.

\textbf{Adversarial Loss}:
for each generator and discriminator pair, the \textit{adversarial loss} is defined as follows.
\if\mycmd1
\begin{equation}
\label{eq:adversarial_loss}
    \mathcal{L}_\text{adv}(G, D_Y, X, Y) = \mathbb{E}_{x \sim p_\text{data}(x)}[\text{log}(1-D_Y(G(x)))] + \mathbb{E}_{y \sim p_\text{data}(y)}[\text{log}(D_Y(y))],
\end{equation}
\else
\begin{align}
\label{eq:adversarial_loss}
    \mathcal{L}_\text{adv}(G, D_Y, X, Y) &= \mathbb{E}_{x \sim p_\text{data}(x)}[\text{log}(1-D_Y(G(x)))] \nonumber \\
    &+ \mathbb{E}_{y \sim p_\text{data}(y)}[\text{log}(D_Y(y))],
\end{align}
\fi
where $G(x)$ is a generated image by a generator $G$.
$G$ tries to make $D_Y$ as difficult as possible to distinguish generated samples $G(x)$ from real samples $y$ whereas $D_Y$ should not be deceived by $G$.
The relationship can be formulated as $\min_G \max_{D_Y} \mathcal{L}_{\text{adv}}(G, D_Y, X, Y)$.
Hence, the generator actually should minimize the following loss.
\if\mycmd1
\begin{align}
    \label{eq:loss_adv_G}
    \mathcal{L}_{\text{adv}, G}(G, D_Y, X, Y) &= \mathbb{E}_{x \sim p_\text{data}(x)}[\text{log}(1-D_Y(G(x)))].
\end{align}
\else
\begin{align}
    \label{eq:loss_adv_G}
    \!\!\!\!\mathcal{L}_{\text{adv}, G}(G, D_Y, X, Y) = \mathbb{E}_{x \sim p_\text{data}(x)}[\text{log}(1\!-\!D_Y(G(x)))].
\end{align}
\fi
For other pairs, ($F$, $D_X$) and ($G^\text{s}$, $D_X^\text{s}$), the adversarial loss can be obtained by replacing ($G$, $D_Y$) in (\ref{eq:adversarial_loss}) with ($F$, $D_X$) and ($G^\text{s}$, $D_X^\text{s}$).
In addition, in (\ref{eq:adversarial_loss}), $X$ and $Y$ are replaced with $Y$ and $X$.

\textbf{Cycle-Consistency Loss}: the \textit{cycle-consistency loss} \cite{Zhu_17} is defined as
\begin{equation}
    \mathcal{L}_\text{cyc}(G, F) = \mathcal{L}_{ \text{cyc}, G}(F) + \mathcal{L}_{\text{cyc}, F}(G),
\end{equation}
where $\mathcal{L}_{G, \text{cyc}}(F)$ and $\mathcal{L}_{F, \text{cyc}}(G)$ is a cycle-consistency loss for each generator, which is defined as
\begin{align}
    \mathcal{L}_{\text{cyc}, G}(F) &= \mathbb{E}_{x \sim p_\text{data}(x)}[\lVert F(G(x)) - x \rVert_1], \\
    \mathcal{L}_{\text{cyc}, F}(G) &= \mathbb{E}_{y \sim p_\text{data}(y)}[\lVert G(F(y)) - y \rVert_1].
\end{align}
This cycle-consistency loss is used to induce a sample $x_i$ to be mapped to a desired sample $y_j$.
Note that the adversarial loss guarantees that via a learned mapping function samples in a domain $X$ are mapped to samples in a domain $Y$, but the learned mapping function cannot translate a sample $x_i$ to a intended $y_j$.
In other words, the adversarial loss can guarantee translation between data distributions.
Hence, to obtain a mapping function between individual samples, the cycle-consistency loss should be used in the training procedure for generators.

\textbf{Multi-Scale Discriminators' Feature Loss}:
multiple discriminators are utilized to distinguish between a sample $y_j$ and a synthesized output $G(x_i)$ \cite{Park_19}.
$M$ Discriminators are trained to distinguish $y_j$ and $G(x_i)$ at $M$ different scales, which allows each discriminator to examine $y_j$ and $G(x_i)$ at a different view.
As the size of $y_j$ and $G(x_i)$ becomes smaller, a discriminator has a wider view of $y_j$ and $G(x_i)$ since receptive field sizes of all the discriminators are the same. 
By using multiple discriminators, for a generator $G^\text{s}$, a GAN feature matching loss is defined as follows.
\if\mycmd1
\begin{equation}
\label{eq:loss_FM_G}
    \mathcal{L}_{\text{FM}, G^\text{s}}(D_{X,1}^\text{s}, \ldots, D_{X,M}^\text{s}) = \sum_{k=1}^{M} \frac{1}{M} \mathbb{E}_{(y,x)} \Bigl [ \sum_{i=1}^{T} \frac{1}{N_i} \lVert D_{X,k}^{\text{s}, i}(x) - D_{X,k}^{\text{s}, i}(G^\text{s}(y)) \rVert_{1} \Bigr ],
\end{equation}
\else
\begin{align}
\label{eq:loss_FM_G}
    \!\!\!\!&\mathcal{L}_{\text{FM}, G^\text{s}}(D_{X,1}^\text{s}, \ldots, D_{X,M}^\text{s}) \nonumber \\
    \!\!\!\!&= \!\! \sum_{k=1}^{M} \!\frac{1}{M} \mathbb{E}_{(y,x)} \!\Bigl [ \sum_{i=1}^{T} \frac{1}{N_i} \lVert D_{X,k}^{\text{s}, i}(x) \!-\! D_{X,k}^{\text{s}, i}(G^\text{s}(y)) \rVert_{1} \! \Bigr ],
\end{align}
\fi
where $\mathbb{E}_{(y, x) \sim p_\text{data}(y, x)} \triangleq \mathbb{E}_{(y,x)}$ for simplicity.
$M$ is the number of discriminators, $D_{X, k}^\text{s}(\cdot)$ is the $M$-th discriminator, and $D_{X, k}^{\text{s}, i}$ is denoted as the $i$-th layer feature extractor of $D_{X, k}^\text{s}(\cdot)$.
$N_i$ means the number of elements in each layer, and $T$ is the number of feature layers.
Note that $G^\text{s}$ can learn how to translate a semantic image to a photorealistic image at both coarse and fine views since discriminators distinguish $x$ and $G^\text{s}(y)$ at $M$ different views.

\textbf{VGG Perceptual Loss}: the work \cite{Park_19} utilizes the perceptual loss in \cite{Johnson_16}.
The VGG perceptual loss is obtained by the 19-layer VGG network \cite{vgg}.
The VGG perceptual loss is defined as
\if\mycmd1
\begin{equation}
\label{eq:loss_VGG_G}
    \mathcal{L}_\text{VGG}(\psi, x, G^\text{s}(y))=\sum_{i \in S_\text{I}} \frac{1}{C_i H_i W_i}\lVert \psi_i(G^\text{s}(y)) - \psi_i(x) \rVert_{1},
\end{equation}
\else
\begin{equation}
\label{eq:loss_VGG_G}
    \mathcal{L}_\text{VGG}(\psi, x, G^\text{s}(y))=\sum_{i \in S_\text{I}} \frac{\lVert \psi_i(G^\text{s}(y)) - \psi_i(x) \rVert_{1}}{C_i H_i W_i},
\end{equation}
\fi
where $S_\text{I}$ is the set including VGG's layer indexes, $\psi$ is the 19-layer VGG network and $\psi_i$ is denoted as the $i$-th layer of $\psi$.
For $\psi_i$, $C_i$, $H_i$, and $W_i$ are the number of channels, the height, and the width, respectively.
By minimizing $\mathcal{L}_\text{VGG}(\cdot)$, $G^\text{s}$ can generate a photorealistic image $G^\text{s}(y)$, visually indistinguishable from $x$ in the feature-level perspective.

\subsection{Our Modifications for Anonymization System}
\subsubsection{Modification on Segmentation Generator's Loss}
The goal of a segmentation generator $G$ is to generate semantic images with which a synthesis generator $G^\text{s}$ makes well-synthesized images.

\textbf{Modification}: to consider the performance of $G^\text{s}$ in the loss of $G$, we define the loss of $G$ as follows.
\begin{align}
\label{eq:loss_G_segmentation}
    &\mathcal{L}_{G}(G, F, D_Y, G^\text{s}) \nonumber\\
    &=  \underbrace{\mathcal{L}_\text{\text{adv}, G}(G, D_Y, X, Y) + \lambda_\text{cyc} \mathcal{L}_{\text{cyc}, G}(F)}_{\text{the original loss of $G$}} \nonumber \\ 
    & + 
    \underbrace{\lambda_\text{s} \mathcal{L}_{G^\text{s}}(G^\text{s}, D_{X, 1}^\text{s}, \ldots ,D_{X, M}^\text{s}, G) + \lambda_\text{dist} \mathcal{L}_\text{dist}(G)}_{\text{the newly added term}},
\end{align}
where $\mathcal{L}_\text{dist}(G) = \mathbb{E}_{x \sim p_\text{data}(x)}[\lVert G(x)-y \rVert_{1}]$, and $\mathcal{L}_\text{dist}(G)$ could further reduce the space of possible mapping functions with $\mathcal{L}_{\text{cyc}, G}(F)$.
$\mathcal{L}_{G^\text{s}}$ is the loss of a synthesis generator $G^\text{s}$, and will be explained in Section~\ref{sec:Modifications_on_Synthesis} in detail.
In addition, $\lambda_\text{cyc}, \lambda_\text{s}$, and $\lambda_\text{dist}$ control the relative importance of each loss.

By adding $\mathcal{L}_{G^\text{s}}(G^\text{s}, D_{X, 1}^\text{s}, \ldots ,D_{X, M}^\text{s}, G)$, the generator $G$ will be trained to generate a semantic image minimizing the loss of $G^\text{s}$.

\textbf{Objective of Segmentation Learning Part}:
Hence, the full objective of segmentation-learning part is defined as:
\if\mycmd1
\begin{align}
\label{eq:objective_segmentation}
    \mathcal{L}_\text{seg}(G, F, D_X, D_Y, G^\text{s}) &= \mathcal{L}_{G}(G, F, D_Y, G^\text{s}) + \mathbb{E}_{y\sim p_\text{data}(y)}[D_Y(y)]
    \nonumber \\
    &+ \mathcal{L}_{F}(G, F, D_X) +\mathbb{E}_{x\sim p_\text{data}(x)}[D_X(x)],
\end{align}
\else
\begin{align}
\label{eq:objective_segmentation}
    &\mathcal{L}_\text{seg}(G, F, D_X, D_Y, G^\text{s}) \nonumber \\
    &= \mathcal{L}_{G}(G, F, D_Y, G^\text{s}) + \mathbb{E}_{y\sim p_\text{data}(y)}[D_Y(y)]
    \nonumber \\
    &+ \mathcal{L}_{F}(G, F, D_X) +\mathbb{E}_{x\sim p_\text{data}(x)}[D_X(x)],
\end{align}
\fi
where $\mathcal{L}_{F}(G,\! F,\! D_X) \!\!=\!\! \mathcal{L}_{\text{adv}, F}(F, \!D_X, \!Y, \!X) + \lambda_\text{cyc} \mathcal{L}_{\text{cyc}, F}(G)$.

The segmentation learning part will solve (\ref{eq:objective_segmentation}) as follows:
\begin{equation}
\label{eq:solve_segmentation_objective}
    G^{*}, F^{*} = \argmin_{G, F} \max_{D_X, D_Y} \mathcal{L}_\text{seg}(G, F, D_X, D_Y, G^\text{s}).
\end{equation}

\subsubsection{Modifications on Synthesis Generator's Loss}
\label{sec:Modifications_on_Synthesis}
There are two main challenges to hinder the learning of synthesis generator to anonymize faces.
We introduce loss of a synthesis generator $G^\text{s}$, and then explain each challenge. 
To obtain a synthesis generator to achieve our system purpose, we have a modification on the loss of $G^\text{s}$ and the loss of $D_{X,k}^\text{s}, \forall k$.
In addition, we modify the way to train the discriminator $D_{X,k}^\text{s}$.

In \cite{Park_19}, by using (\ref{eq:loss_adv_G}), (\ref{eq:loss_FM_G}), and (\ref{eq:loss_VGG_G}), the loss of a synthesis generator can be written as
\if\mycmd1
\begin{align}
\label{eq:loss_G_synthesis}
    \mathcal{L}_{G^\text{s}}(G^\text{s}, D_{X,1}^\text{s}, \ldots D_{X,M}^\text{s}, G) &= \sum_{k=1}^{M} \frac{1}{M} \Bigl \{ \mathcal{L}_{\text{adv}, G^\text{s}}(G^\text{s}, D_{X, k}^\text{s}, Y, X, G(x)) + \mathcal{L}_{\text{FM}, G^\text{s}}(D_{X,k}^\text{s})  \Bigr \} \nonumber \\ 
    &+ \mathcal{L}_\text{VGG}(\psi, x, G^\text{s}(G(x))),
\end{align}
\else
\begin{align}
\label{eq:loss_G_synthesis}
    &\mathcal{L}_{G^\text{s}}(G^\text{s}, D_{X,1}^\text{s}, \ldots D_{X,M}^\text{s}, G) \nonumber \\
    &=\!\! \sum_{k=1}^{M} \! \frac{1}{M} \Bigl \{ \! \mathcal{L}_{\text{adv}, G^\text{s}}(G^\text{s}, D_{X, k}^\text{s}, Y, X, G(x)) + \mathcal{L}_{\text{FM}, G^\text{s}}(D_{X,k}^\text{s}) \!  \Bigr \} \nonumber \\ 
    &+ \mathcal{L}_\text{VGG}(\psi, x, G^\text{s}(G(x))),
\end{align}
\fi
where we introduce for simplicity $\mathcal{L}_{\text{FM}, G^\text{s}}(D_{X,k}^\text{s})=\mathbb{E}_{(y,x)} \Bigl [ \sum_{i=1}^{T} \frac{1}{N_i} \lVert D_{X,k}^{\text{s}, i}(x) - D_{X,k}^{\text{s}, i}(G^\text{s}(y)) \rVert_{1} \Bigr ]$ in (\ref{eq:loss_FM_G}).
In addition, $\mathcal{L}_{\text{adv}, G^\text{s}}(G^\text{s}, D_{X, k}^\text{s}, Y, X, G(x))$ is redefined as
\if\mycmd1
\begin{equation}
\label{eq:loss_synthesis_G_redefined}
    \mathcal{L}_{\text{adv}, G^\text{s}}(G^\text{s}, D_{X, k}^\text{s}, Y, X, G(x)) = \mathbb{E}_{y\sim p_\text{data}(y)} \Bigl [\text{log}(1-D_{X,k}^\text{s}(G^\text{s}(G(x)))) \Bigr ].
\end{equation}
\else
\begin{align}
\label{eq:loss_synthesis_G_redefined}
    &\mathcal{L}_{\text{adv}, G^\text{s}}(G^\text{s}, D_{X, k}^\text{s}, Y, X, G(x)) \nonumber \\
    &= \mathbb{E}_{\hat{y}\sim p_\text{data}(y)} \Bigl [\text{log}(1-D_{X,k}^\text{s}(G^\text{s}(\hat{y}))) \Bigr ],
\end{align}
\fi
where $\hat{y}=G(x)$.

\textbf{Challenge in VGG Perceptual Loss}:
in the synthesis-learning part, the loss (\ref{eq:loss_G_synthesis}) should be minimized to train the generator $G^\text{s}$.
The minimization leads to reduce $\mathcal{L}_{\text{VGG}}(\psi, x, G^\text{s}(G(x)))$, and thus the distance between features of $x$ and $G^\text{s}(G(x))$ is also reduced during the training of $G^\text{s}$.
As a result, the generator $G^\text{s}$ is trained to generate a photorealistic image $G^\text{s}(G(x))$ that can be almost the same as the original photorealistic image $x$.
This trained generator cannot be utilized for our face-anonymizing system.

\textbf{Modification on VGG Perceptual Loss}:
to prevent the distance between $G^\text{s}(G(x))$ and $x$ from being reduced to a very small value, we introduce margins to the VGG perceptual loss (\ref{eq:loss_VGG_G}) as follows.
\if\mycmd1
\begin{equation}
\label{eq:loss_VGG_G_with_margin}
    \mathcal{L}_\text{VGG}(\psi, x, G^\text{s}(y), S_\text{I})=\sum_{i \in S_\text{I}} \max \Bigl (0,  \frac{1}{C_i H_i W_i} \lVert \psi_i(G^\text{s}(y)) - \psi_i(x) \rVert_{1} - \epsilon_{m(i)} \Bigr),
\end{equation}
\else
\begin{align}
\label{eq:loss_VGG_G_with_margin}
    &\mathcal{L}_\text{VGG}(\psi, x, G^\text{s}(y), S_\text{I})\nonumber \\
    &=\sum_{i \in S_\text{I}} \max \Bigl (0,  \frac{\lVert \psi_i(G^\text{s}(y)) - \psi_i(x) \rVert_{1}}{C_i H_i W_i} - \epsilon_{m(i)} \Bigr),
\end{align}
\fi
where $\Upsilon = \{ \epsilon_1, \ldots, \epsilon_{|S_\text{I}|} \}$ and $|S_\text{I}|$ is the number of elements in $S_\text{I}$.
$m(i)$ is a mapping function to find for $S_\text{I}$ a VGG's layer index corresponding to $i$.

Note that a margin value allows the distance between the $i$-th VGG layers for $x$ and $G^\text{s}(G(x))$ to be at least $\epsilon_i$ value.
Hence, a photorealistic image $G^\text{s}(G(x))$ can have different features from features of the original image $x$, which could make $G^\text{s}(G(x))$ look different from $x$.

\textbf{Challenge in Adversarial Loss and Multi-Scale Discriminators' Feature Loss}:
to explain our additional modifications, we need to comprehend about how a discriminator $D_{X, k}^\text{s}$ works.
Based on the understanding, we describe a hindrance to the learning of our synthesis generator $G^\text{s}$.
In addition, we modify the adversarial losses of $G^\text{s}$ and $D_{X, k}^\text{s}$, and the multi-scale discriminators' feature loss of $G^\text{s}$.

To minimize (\ref{eq:loss_synthesis_G_redefined}), $G^\text{s}$ should make a synthesized face $G^\text{s}(G(x))$ look like a face in the training dataset, and thus tends to translate $G(x)$ to $x$ that is what our system should anonymize.
Specifically, in (\ref{eq:loss_synthesis_G_redefined}), a discriminator $D_{X,k}^\text{s}$ examines $G^\text{s}(G(x))$ to determine if $G^\text{s}(G(x))$ is from the training dataset $X$ or is arbitrarily generated.
The generated image $G^\text{s}(G(x))$ contains an entire face, and thus $D_{X,k}^\text{s}$ is trained to determine whether the entire face in $G^\text{s}(G(x))$ is from the training dataset.
As a result, to deceive $D_{X,k}^\text{s}$, $G^\text{s}$ is trained to generate $x$ from $G(x)$.

\textbf{Modifications on Adversarial Loss and Multi-Scale Discriminators' Feature Loss}:
to prevent that $G^\text{s}$ regenerates the almost same face as $x$, we have modification on the adversarial losses of $G^\text{s}$ and $D_{X,k}^\text{s}$.
The reason that $G^\text{s}$ reproduces $x$ is because a discriminator $D_{X,k}^\text{s}$ examines if the entire face in $G^\text{s}(G(x))$ is from a training dataset including $x$.
In other words, to deceive $D_{X,k}^\text{s}$ checking an entire face, $G^\text{s}$ necessarily makes a face from the domain $X$, which greatly reduces the space of possible mapping.
In addition, $G^\text{s}$ should produce a photorealistic face by maintaining the shape and location of each facial part in a semantic-face image, which also further reduces the space of possible mapping.

For expanding the space of possible mapping, we limit a discriminator $D_{X,k}^\text{s}$ to investigate each facial component, not entire face.
By applying the idea, the adversarial loss is rewritten as follows.
\if\mycmd1
\begin{align}
    \label{eq:loss_adv_synthesis_modified}
    \mathcal{L}_\text{adv}^{s}(G^\text{s}, D_{X,1}^\text{s}, \ldots, D_{X,M}^\text{s}, Y, X, S_\xi) &= 
    \sum_{k=1}^{M} \frac{1}{M} \Bigl \{ \mathbb{E}_{y\sim p_\text{data}(y)} \Bigl [ \sum_{i\in S_{\xi}} \frac{1}{|S_{\xi}|} \text{log}(1-D_{X,k}^\text{s}(\xi_{i}(G^\text{s}(\hat{y})))) \Bigr ] \Bigr \} \nonumber \\
    &+
    \sum_{k=1}^{M} \frac{1}{M} \Bigl \{ \mathbb{E}_{x\sim p_\text{data}(x)} \Bigl [  \sum_{i\in S_{\xi}} \frac{1}{|S_{\xi}|} \text{log}(D_{X,k}^\text{s}(\xi_i(x))) \Bigr ] \Bigr \},
\end{align}
\else
\begin{align}
    \label{eq:loss_adv_synthesis_modified}
    &\mathcal{L}_\text{adv}^{s}(G^\text{s}, D_{X,1}^\text{s}, \ldots, D_{X,M}^\text{s}, Y, X, S_\xi) \nonumber \\
    &= 
    \sum_{k=1}^{M}\! \frac{1}{M} \! \Bigl \{ \mathbb{E}_{y\sim p_\text{data}(y)} \Bigl [ \sum_{i\in S_{\xi}} \!\!\frac{1}{|S_{\xi}|} \text{log}(1\!-\!D_{X,k}^\text{s}(\xi_{i}(G^\text{s}(\hat{y})))) \Bigr ] \! \Bigr \} \nonumber \\
    &+
    \sum_{k=1}^{M} \frac{1}{M} \Bigl \{ \mathbb{E}_{x\sim p_\text{data}(x)} \Bigl [  \sum_{i\in S_{\xi}} \frac{1}{|S_{\xi}|} \text{log}(D_{X,k}^\text{s}(\xi_i(x))) \Bigr ] \Bigr \},
\end{align}
\fi
where we denote the output of our segmentation generator as $\hat{y}=G(x)$, $\xi_i$ is the extractor to extract pixels corresponding to the label index $i$, $S_{\xi}$ is the set including extracted labels' index, and $|S_{\xi}|$ is the number of elements in $S_\xi$.
For example, if $i=2$ and the label index 2 indicates a nose in a face, all pixels in $\xi_2(G^\text{s}(\hat{y}))$ become zero except for the pixels corresponding to the nose.

According to (\ref{eq:loss_adv_synthesis_modified}), our modification allows a discriminator $D_{X,k}^\text{s}$ to examine a part of a face instead of observing all facial parts at a time.
This approach allows discriminators to learn the distribution of each facial component instead of learning the distribution of an entire face.
Hence, a discriminator tries to distinguish each part of a face in $G^\text{s}(\hat{y})$ from that of a face in $x$, which could widen the space of possible mapping in the entire face's point of view.
In addition, by setting $|S_\xi|<N_\text{f}$ where $N_\text{f}$ denotes the number of labels in a face, we make discriminators observe some parts of an entire face, and thus the generator $G^\text{s}$ could have wider space of possible mapping for the other parts not examined by discriminators.

\if\mycmd1
In the same vein, the multi-scale discriminators' feature loss can be also redefined as
\begin{equation}
\label{eq:loss_multi_scale_disc_feature_modified}
    \mathcal{L}_{\text{FM}, G^\text{s}}(D_{X,1}^\text{s}, \ldots, D_{X,M}^\text{s}, S_\xi) = 
    \sum_{k=1}^{M} \frac{1}{M} 
    \underbrace{\mathbb{E}_{(y,x)} \Bigl [ \sum_{i=1}^{T} \frac{1}{N_i} 
    \sum_{j\in S_\xi} \frac{1}{|S_\xi|} \lVert D_{X,k}^{\text{s}, i}(\xi_j(x)) - D_{X,k}^{\text{s}, i}(\xi_j(G^\text{s}(\hat{y}))) \rVert_{1} \Bigr ]}_{\mathcal{L}_{\text{FM},G^\text{s}}(D_{X,k}^\text{s}, S_\xi)}.
\end{equation}
\else
In the same vein, the multi-scale discriminators' feature loss can be also redefined as (\ref{eq:loss_multi_scale_disc_feature_modified}).
\begin{figure*}
\begin{align}
\label{eq:loss_multi_scale_disc_feature_modified}
    &\mathcal{L}_{\text{FM}, G^\text{s}}(D_{X,1}^\text{s}, \ldots, D_{X,M}^\text{s}, S_\xi) = 
    \sum_{k=1}^{M} \frac{1}{M} 
    \underbrace{\mathbb{E}_{(y,x)} \Bigl [ \sum_{i=1}^{T} \frac{1}{N_i} 
    \sum_{j\in S_\xi} \frac{1}{|S_\xi|} \lVert D_{X,k}^{\text{s}, i}(\xi_j(x)) D_{X,k}^{\text{s}, i}(\xi_j(G^\text{s}(\hat{y}))) \rVert_{1} \Bigr ]}_{\mathcal{L}_{\text{FM},G^\text{s}}(D_{X,k}^\text{s}, S_\xi)}.
\end{align}
\end{figure*}
\fi


In spite of our modifications on $\mathcal{L}_\text{adv}^\text{s}(\cdot)$ and $\mathcal{L}_{\text{FM}, G^\text{s}}(\cdot)$, there is still room for $G^\text{s}$ to learn to regenerate $x$ since $\mathcal{L}_{\text{FM},G^\text{s}}(D_{X,k}^\text{s}, S_\xi)$ still compares features of $G^\text{s}(G(x))$ and $x$, which are extracted by $D_{X,k}^\text{s}$.
\if\mycmd1
Hence, we slightly modify (\ref{eq:loss_multi_scale_disc_feature_modified}) as follows.
\begin{align}
\label{eq:loss_multi_scale_disc_feature_modified_2}
    \mathcal{L}_{\text{FM}, G^\text{s}}(D_{X,1}^\text{s}, \ldots, D_{X,M}^\text{s}, S_\xi) = 
    \sum_{k=1}^{M} \frac{1}{M} 
    \underbrace{\mathbb{E}_{(y,x,\tilde{x})} \Bigl [ \sum_{i=1}^{T} \frac{1}{N_i} 
    \sum_{j\in S_\xi} \frac{1}{|S_\xi|} \lVert D_{X,k}^{\text{s}, i}(\xi_j(\tilde{x})) - D_{X,k}^{\text{s}, i}(\xi_j(G^\text{s}(\hat{y}))) \rVert_{1} \Bigr ]}_{\mathcal{L}_{\text{FM},G^\text{s}}(D_{X,k}^\text{s}, S_\xi)},
\end{align}
where $\tilde{x} \neq x$ but $\tilde{x}$ is in the same training dataset of $x$.
By the modification, $\mathcal{L}_{\text{FM},G^\text{s}}(D_{X,k}^\text{s}, S_\xi)$ compares features of $G^\text{s}(G(x))$ to features of $\tilde{x}$, which can help $G\text{s}$ learning to generate a different face from $x$.
\else
Hence, we slightly modify (\ref{eq:loss_multi_scale_disc_feature_modified}) to (\ref{eq:loss_multi_scale_disc_feature_modified_2}).
\begin{figure*}
\begin{align}
\label{eq:loss_multi_scale_disc_feature_modified_2}
    \mathcal{L}_{\text{FM}, G^\text{s}}(D_{X,1}^\text{s}, \ldots, D_{X,M}^\text{s}, S_\xi) = 
    \sum_{k=1}^{M} \frac{1}{M} 
    \underbrace{\mathbb{E}_{(y,x,\tilde{x})} \Bigl [ \sum_{i=1}^{T} \frac{1}{N_i} 
    \sum_{j\in S_\xi} \frac{1}{|S_\xi|} \lVert D_{X,k}^{\text{s}, i}(\xi_j(\textcolor{blue}{\tilde{x}})) - D_{X,k}^{\text{s}, i}(\xi_j(G^\text{s}(\hat{y}))) \rVert_{1} \Bigr ]}_{\mathcal{L}_{\text{FM},G^\text{s}}(D_{X,k}^\text{s}, S_\xi)},
\end{align}
\noindent\makebox[\linewidth]{\rule{17cm}{0.4pt}}
\end{figure*}
In (\ref{eq:loss_multi_scale_disc_feature_modified_2}), $\mathbb{E}_{(x,y,\tilde{x})\sim p_\text{data}(x, y, x)} \triangleq \mathbb{E}_{x,y,\tilde{x}}$, and $\tilde{x} \neq x$ but $\tilde{x}$ is from the same training dataset of $x$.
By the modification, $\mathcal{L}_{\text{FM},G^\text{s}}(D_{X,k}^\text{s}, S_\xi)$ compares features of $G^\text{s}(G(x))$ to features of $\tilde{x}$, which can help $G^\text{s}$ learning to generate a different face from $x$.
\fi

\textbf{Objective of Synthesis Learning Part}: Based on (\ref{eq:loss_VGG_G_with_margin}), (\ref{eq:loss_adv_synthesis_modified}), and (\ref{eq:loss_multi_scale_disc_feature_modified_2}), our full objective of synthesis-learning part is defined as:
\if\mycmd1
\begin{align}
\label{eq:objective_synthesis_learning}
    \mathcal{L}_\text{syn}(G^\text{s}, D_{X,1}^\text{s}, \ldots, D_{X,M}^\text{s}, S_\xi, G) &= \mathcal{L}_\text{adv}^\text{s}(G^\text{s}, D_{X,1}^\text{s}, \ldots, D_{X,M}^\text{s}, Y, X, S_\xi) 
    + \mathcal{L}_{\text{FM},  G^\text{s}}(D_{X,1}^\text{s}, \ldots, D_{X,M}^\text{s}, S_\xi) \nonumber \\
    &+
    \mathcal{L}_\text{VGG}(\psi, x, G^\text{s}(y), S_\text{I}) +
    \mathcal{L}_{\text{cyc}, G^\text{s}}(G),
\end{align}
\else
\begin{align}
\label{eq:objective_synthesis_learning}
    &\mathcal{L}_\text{syn}(G^\text{s}, D_{X,1}^\text{s}, \ldots, D_{X,M}^\text{s}, S_\xi, G) \nonumber \\ 
    &= \mathcal{L}_\text{adv}^\text{s}(G^\text{s}, D_{X,1}^\text{s}, \ldots, D_{X,M}^\text{s}, Y, X, S_\xi) \nonumber \\
    &+ \mathcal{L}_{\text{FM},  G^\text{s}}(D_{X,1}^\text{s}, \ldots, D_{X,M}^\text{s}, S_\xi) \nonumber \\
    &+
    \mathcal{L}_\text{VGG}(\psi, x, G^\text{s}(y), S_\text{I}) +
    \mathcal{L}_{\text{cyc}, G^\text{s}}(G),
\end{align}
\fi
where $\mathcal{L}_{\text{cyc}, G^\text{s}}(G) = \mathbf{E}_{x \sim p_\text{data}(x)}[\lVert G(G^\text{s}(G(x))) - G(x) \rVert_{1}]$ that allows a synthesized image $G^\text{s}(G(x))$ to maintain the shape and location of each facial part in $x$.
Since (\ref{eq:loss_multi_scale_disc_feature_modified_2}) compares features of $G^\text{s}(G(x))$ to those of $\tilde{x}$, the generator $G^\text{s}$ can make a synthesized image to remain the shape and location of each facial part in $\tilde{x}$ not in $x$.
Hence, $\mathcal{L}_{\text{cyc}, G^\text{s}}(G)$ helps $G^\text{s}$ to generate synthesized images to remain the shape and location of facial parts in $x$.
Hence, by $\mathcal{L}_{\text{cyc}, G^\text{s}}(G)$ and (\ref{eq:loss_multi_scale_disc_feature_modified_2}), $G^\text{s}$ can generate a synthesized face $G^\text{s}(G(x))$ including facial features of $\tilde{x}$ while maintaining the shape and location of facial components in $x$.

Finally, the synthesis learning part will solve (\ref{eq:objective_synthesis_learning}) as follows:
\begin{equation}
    (G^\text{s})^* = \argmin_{G^\text{s}} \max\limits_{D_{X,k}^\text{s},\forall k} \mathcal{L}_\text{syn}(G^\text{s}, D_{X,1}^\text{s}, \ldots, D_{X,M}^\text{s}, S_\xi, G).
\end{equation}

\subsection{Training Procedure and Details}
\begin{algorithm}[!t]
\caption{Training Procedure for One Epoch}
\label{alg:training_procedure}
\begin{algorithmic}
\Statex \textbf{Output}: Generators, $G^*$ and $(G^\text{s})^*$
\Statex \multiline{\textbf{for} i = 1 : $N_\text{data}$}
\Statex {1) Select $x$, $y$, $\Tilde{x}$ from the dataset $X$ and $Y$}
\Statex {2) Update $G$, $F$, $D_X$, $D_Y$ with $G^\text{s}$}
\Statex {~~~(\ref{eq:objective_segmentation}): $\argmin_{G, F} \max\limits_{D_X, D_Y} \mathcal{L}_\text{seg}(G, F, D_X, D_Y, G^\text{s})$}
\Statex {3) Update $G^\text{s}$, $D_{X,k}^\text{s}, \forall k$ with $G$, $F$, $D_X$, $D_Y$}
\Statex {~~~(\ref{eq:objective_synthesis_learning}): $\argmin_{G^\text{s}} \max\limits_{D_{X,k}^\text{s}, \forall k} \mathcal{L}_\text{syn}(G^\text{s}, D_{X,1}^\text{s}, \ldots, D_{X,M}^\text{s}, S_\xi, G)$}
\end{algorithmic}
\end{algorithm}
\textbf{Procedure}: The overall training procedure is summarized in Algorithm~\ref{alg:training_procedure}, where $N_\text{data}$ is the number of data in the dataset $X$ and $Y$.
By repeating the training procedure, we obtain the optimized $G^*$ and $(G^\text{s})^*$.

\textbf{Details}: In segmentation learning part, for the segmentation generator $G$, we adopt the network architecture in \cite{Johnson_16} that is known for powerful neural-type transfer and use 9 resnet blocks for $256 \times 256$ images, which is applied equally to the generator $F$.
For the discriminators $D_X$ and $D_Y$, we use $70 \times 70$ PatchGANs \cite{Isola_17, Li_16, Ledig_17, Zhu_17}.
In synthesis learning part, we construct our network architecture by applying the Spectrum Norm \cite{Spectrum_Norm} to all the layers in both generator and discriminator.
For our synthesis generator, we use the SPADE generator in \cite{Park_19}.
Finally, we set $M=3$ for our discriminators.

For our training, we set $\lambda_\text{cyc}, \lambda_\text{s}, \lambda_\text{dist}$ to $10$ in (\ref{eq:loss_G_segmentation}).
A solver is set to the ADAM solver \cite{ADAM}  with a batch size of 1.
For the solver, $\beta_1=0.5$ and $\beta_2=0.999$ in segmentation learning part and $\beta_1=0$ and $\beta_2=0.999$ in synthesis learning part.

\textbf{Training Dataset}: We conduct our training procedure with CelebA-HQ dataset \cite{CelebAMask-HQ}.
This dataset contains 30,000 high-resolution face images with 19 semantic classes.
In this work, we modify the semantic dataset by extracting 10 main facial components.
In our modified semantic dataset, semantic classes include skin, nose, eyes, eyebrows, ears, mouth, lip, hair, neck, and eyeglass.
In addition, we utilize a face detector to crop a face in high-resolution face images. 
The reason that cropped face images are needed is that in our system faces in a video frame will be detected by using a face detector.
By conducting this preprocessing, we can train segmentation and synthesis generators, optimized for our system.

Moreover, we create and use various resolution images for a image.
Our segmentation generator requires a specific sized image as input, and thus detected face images should be resized to the specific size.
If the size of a detected face is smaller than the required size, a resized face image is low resolution.
In reality, our system can detect a face in various sizes, various resolutions.
To make our segmentation and synthesis generators work well with various resolutions, we utilize various resolution images for a face image during our training procedure.


\subsection{Face-Anonymizing Algorithm}
\begin{algorithm}[!t]
\caption{Face-Anonymizing Algorithm}
\label{alg:face_anonymizing_algorithm}
\begin{algorithmic}
\Statex \multiline{\textbf{Input}: Face detector $D$; Segmentation generator $G$; \\
Synthesis generator $G^\text{s}$; Video frame $v$}
\Statex \textbf{Output}: Anonymized video frame $\tilde{v}$
\Statex {~1) $D(v) \rightarrow f_\text{D}, N_\text{face}, R_\text{thr}$~~~// Face detection in $v$}
\Statex {~2) \textbf{while} ratio $< R_\text{thr}$ \textbf{do}~~~~// Detect \textit{tightly} a face in $f_\text{D}$}
\Statex {~~~~~~~$D(f_\text{D}) \rightarrow f_\text{D}, R_\text{thr}$}
\Statex {~3) \textbf{if} $N_\text{face} > 0$ \textbf{then} ~~~~~~~~ // Faces exist}
\Statex {~~~~~~~-~$G^\text{s}(G(f_\text{D})) \rightarrow f_\text{A}$ ~~~ // Anonymization of faces}
\Statex \multiline{~~~~~~~-~$f_\text{A}*B^{-1}(G(f_\text{D}))$ \\ 
~~~~~~~~~~~~~$+ f_\text{D}*B(G(f_\text{D})) \rightarrow f_\text{A}$ $\cdots$ \textcircled{1}
}
\Statex {~~~~~~~-~$(v-f_\text{D}) + f_\text{A} \rightarrow \tilde{v}$ ~ // Faces' replacement}
\Statex {~~~~ \textbf{else} ~~~~~~~~~~~~~~~~~~~~~~~ // No faces to anonymize}
\Statex {~~~~~~~~$v \rightarrow \tilde{v}$}
\Statex {~4) Terminate face anonymization in a video frame}
\end{algorithmic}
\end{algorithm}
With the optimized segmentation and synthesis generators, our face-anonymizing procedure is conducted as in Algorithm~\ref{alg:face_anonymizing_algorithm}.
A companion computer conducts Algorithm~\ref{alg:face_anonymizing_algorithm} whenever it receives a video frame.
$f_\text{D}$ denotes detected faces, $N_\text{face}$ is the number of detected faces, $f_\text{A}$ includes all anonymized faces, and $R_\text{thr}$ is a threshold for the ratio of the face size to the image size.
Note that $f_\text{D}$ and $f_\text{A}$ also include the background as well as faces.

\textbf{Maintenance of Background}: In Algorithm~\ref{alg:face_anonymizing_algorithm}, $B(\cdot)$ is a function that makes all nonzero elements in a input semantic image 1.
In a semantic image, zero indicates the background.
$B^{-1}(\cdot)$ is the opposite function of $B(\cdot)$.

Hence, in Algorithm~\ref{alg:face_anonymizing_algorithm}, \textcircled{1} creates a image that mixes the anonymized face in $f_\text{A}$ and the background in $f_\text{D}$, which could preserve the background in the original image.

\textbf{Repetitive Face Detection}: In Algorithm \ref{alg:face_anonymizing_algorithm}, we repeat to conduct the face detector with detected faces until the size of a face in $f_\text{D}$ occupies $R_\text{thr}\times100$ of the size of $f_\text{D}$.
The reason for this repetitive detection is that our generators well anonymize a input image that is full of one face.
\begin{figure}[!t]
\centering
\includegraphics[width=1\columnwidth]{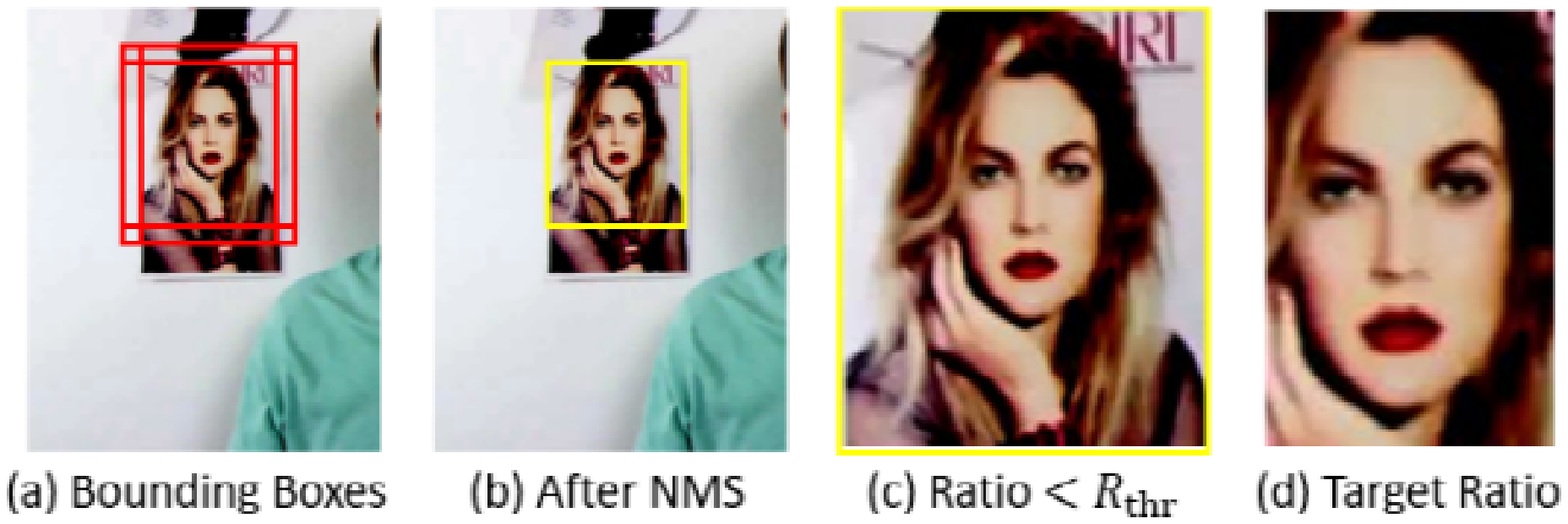}
\caption{Bounding box regression and NMS}
\label{Fig:Repetitive_Detection}
\end{figure}
The face detector finds a face with the bounding box regression and non-maximum suppression (NMS).
The bounding box regression provides several bounding boxes (red boxes) on a face in Fig.~\ref{Fig:Repetitive_Detection}(a).
Then, NMS trims the bounding boxes to obtain a bounding box (the yellow box) in Fig.~\ref{Fig:Repetitive_Detection}(b).
In Fig.~\ref{Fig:Repetitive_Detection}(c), however, the resultant image is not suitable as input for our generators.
Hence, to increase the ratio of the face size to the image size, we repeatedly conduct the face detector on a face until we obtain a image in Fig.~\ref{Fig:Repetitive_Detection}(d).

\subsection{Drone Patrol System}
The drone system consists of the following components: 
(1) Drone control computer operating motors so that the drone can move physically, 
(2) Companion computer conducting face-anonymizing neural networks and performing SLAM, 
(3) Wireless chipset receiving commands from the ground station and transmitting anonymized video frames to it, 
(4) High-resolution camera recording a video;

In our drone, the companion computer is connected with the drone control computer, the wireless chipset, and the high-resolution camera.
We utilize ROS to allow all the components to communicate with each other.
Via wireless communication, the companion computer communicates with the ground station.
The ground station can transmit a command message to the companion computer.
The companion computer sends the received message to drone control computer controls. 
The moving drone continuously records images via the camera, which is passed to our face-anonymzing networks implemented in the companion computer. The anonymized images are sent back to the ground station via the wireless chipset, and thus we can immediately check the results on the screen of the ground station. 
At the same time, the anonymized images are processed by the ORB-SLAM2 algorithm.

\section{Evaluation}
\label{sec:Experiments}
\subsection{Face-Anonymizing Generators Evaluation}

\textbf{Face Detection}: Our system utilizes a lightweight but accurate face detector, called FaceBoxes \cite{zhang2017faceboxes}.
The computing speed is invariant no matter how many faces are in a image, and the accuracy was verified with various face datasets.

\textbf{Test Dataset}: We test our face-anonymizing generators on several datasets.
\begin{itemize}
    \item \textit{CelebA-HQ}: This dataset contains 30,000 high-resolution face images.
    We randomly select 1,500 images for our test. Note that the remaining 28,500 images are used for the training.
    \item \textit{Helen}: This dataset has 2,330 face images \cite{Helen}.
    In addition, this provides 2,330 annotation images to locate 8 facial components, which are skin, eyebrow, eye, nose, lip, inner mouth, and hair.
    \item \textit{Facescrub}: This dataset is large face datset, which contains 106,863 face images of male and female 530 celebrities \cite{Facescrub}.
    \item \textit{FaceForensic}: This dataset provides 1000 video sequences, which has been sourced from 977 youtube videos \cite{roessler2019faceforensicspp}.
     All videos contain a mostly frontal face without occlusions.
\end{itemize}

\begin{figure*}[t]
\centering
\includegraphics[width=0.7\paperwidth]{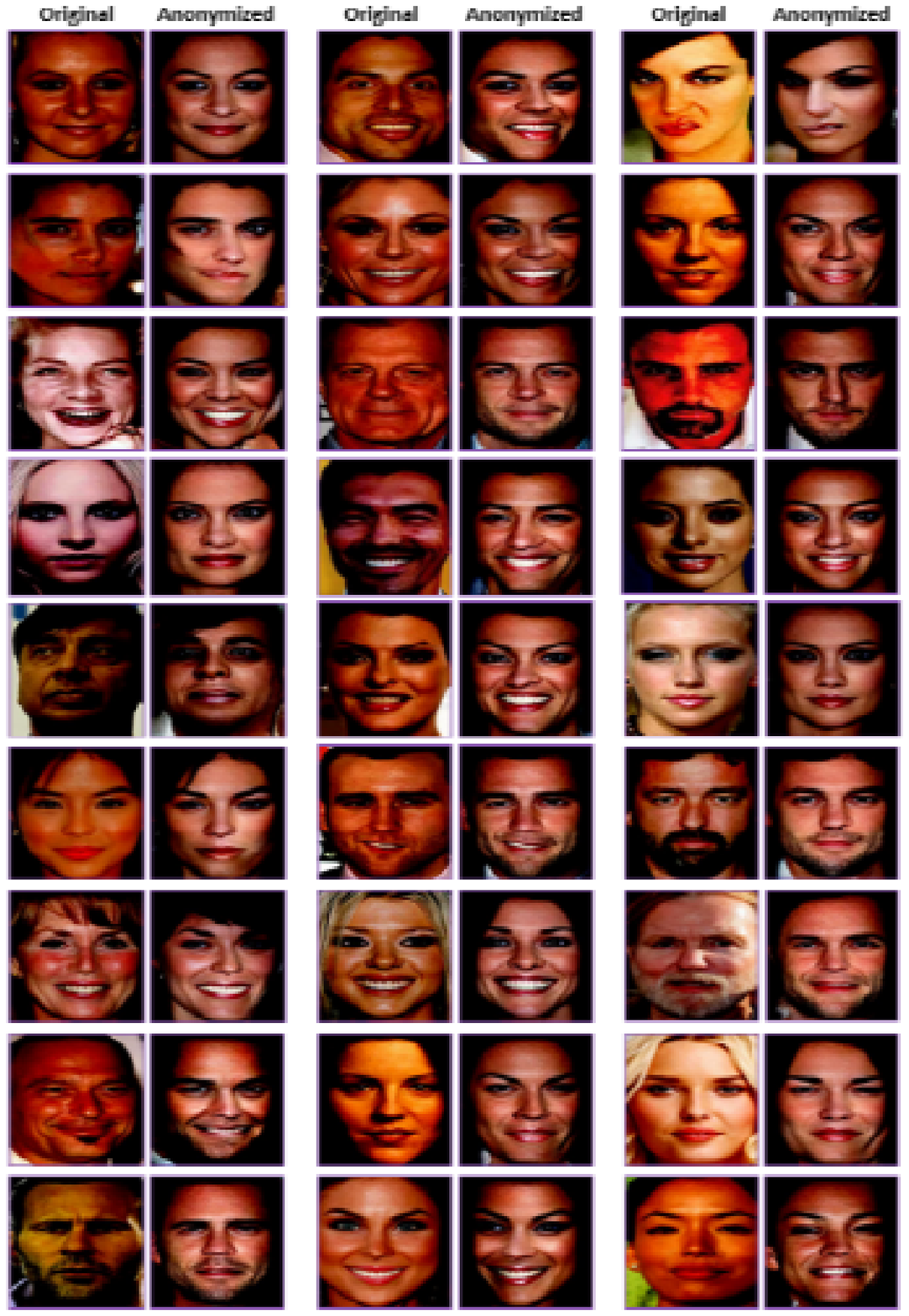}
\caption{Additional Test Results on CelebA-HQ Dataset}
\label{Fig:Result_CelebA-HQ}
\end{figure*}

\begin{figure*}[t]
\centering
\includegraphics[width=0.7\paperwidth]{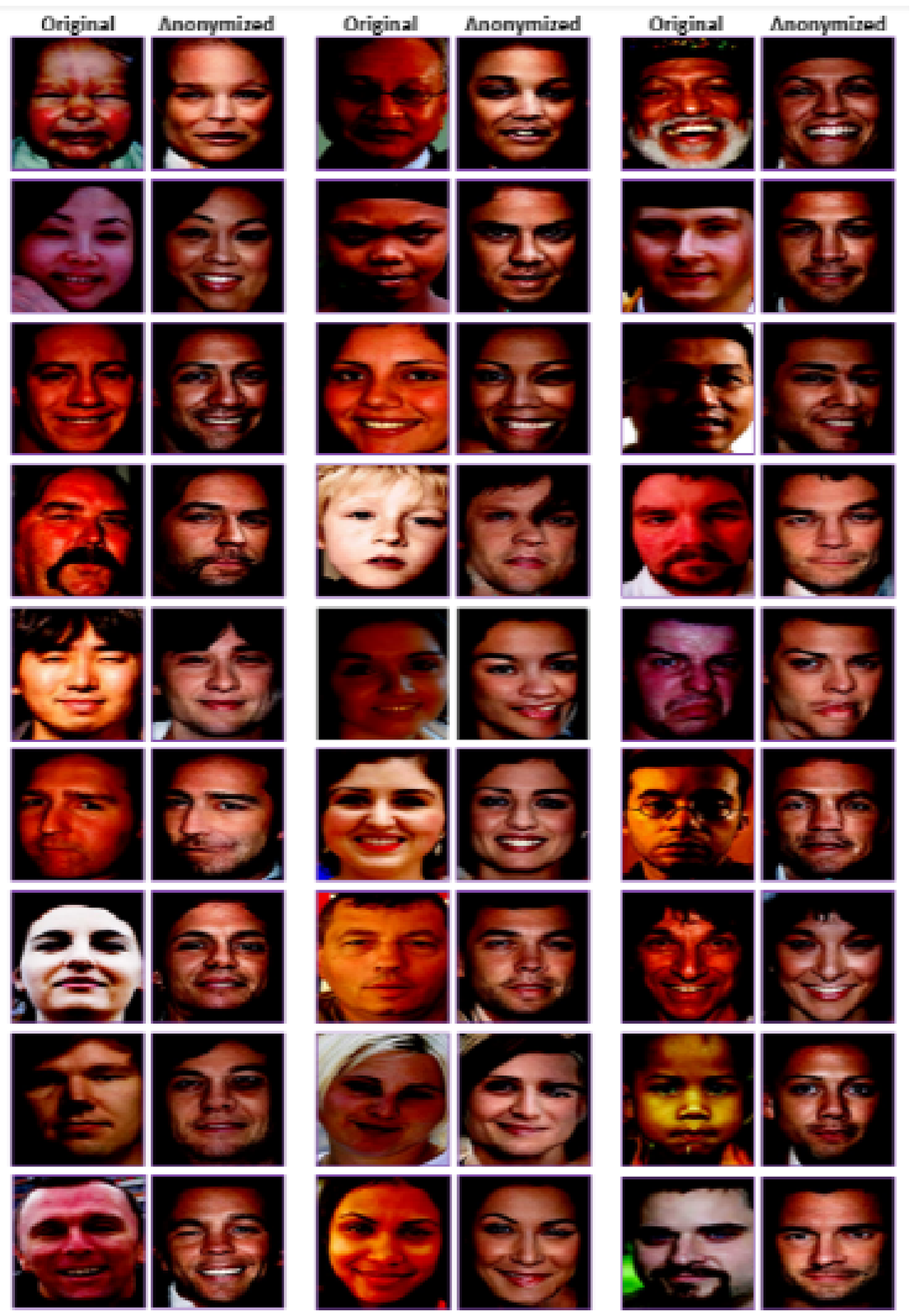}
\caption{Additional Test Results on Helen Dataset}
\label{Fig:Result_Helen}
\end{figure*}

\begin{figure*}[t]
\centering
\includegraphics[width=0.7\paperwidth]{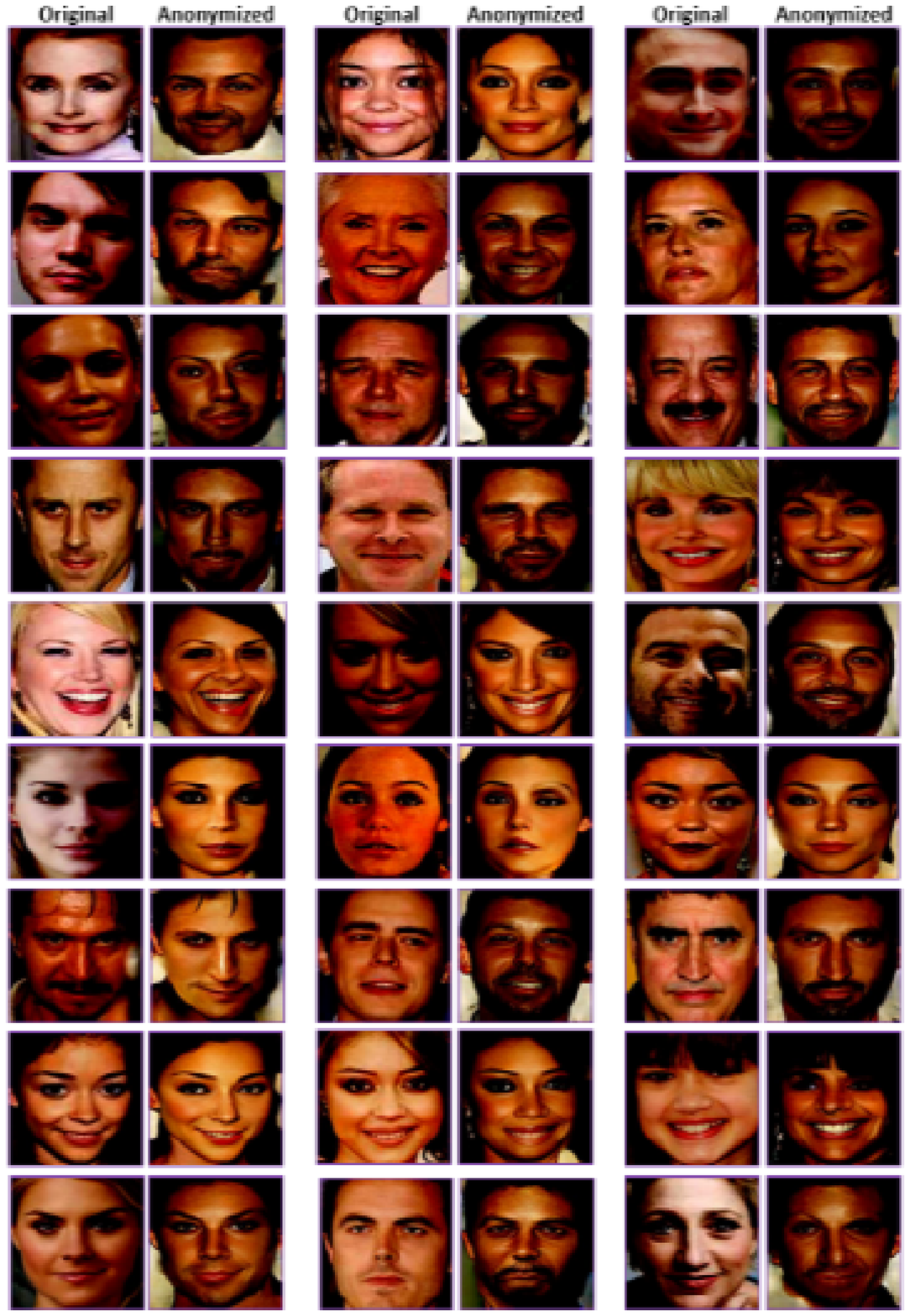}
\caption{Additional Test Results on FaceScrub Dataset}
\label{Fig:Result_FaceScrub}
\end{figure*}

\textbf{Qualitative Evaluation of Face Anonymization}:
\begin{figure}[t]
\centering
\includegraphics[width=0.95\columnwidth]{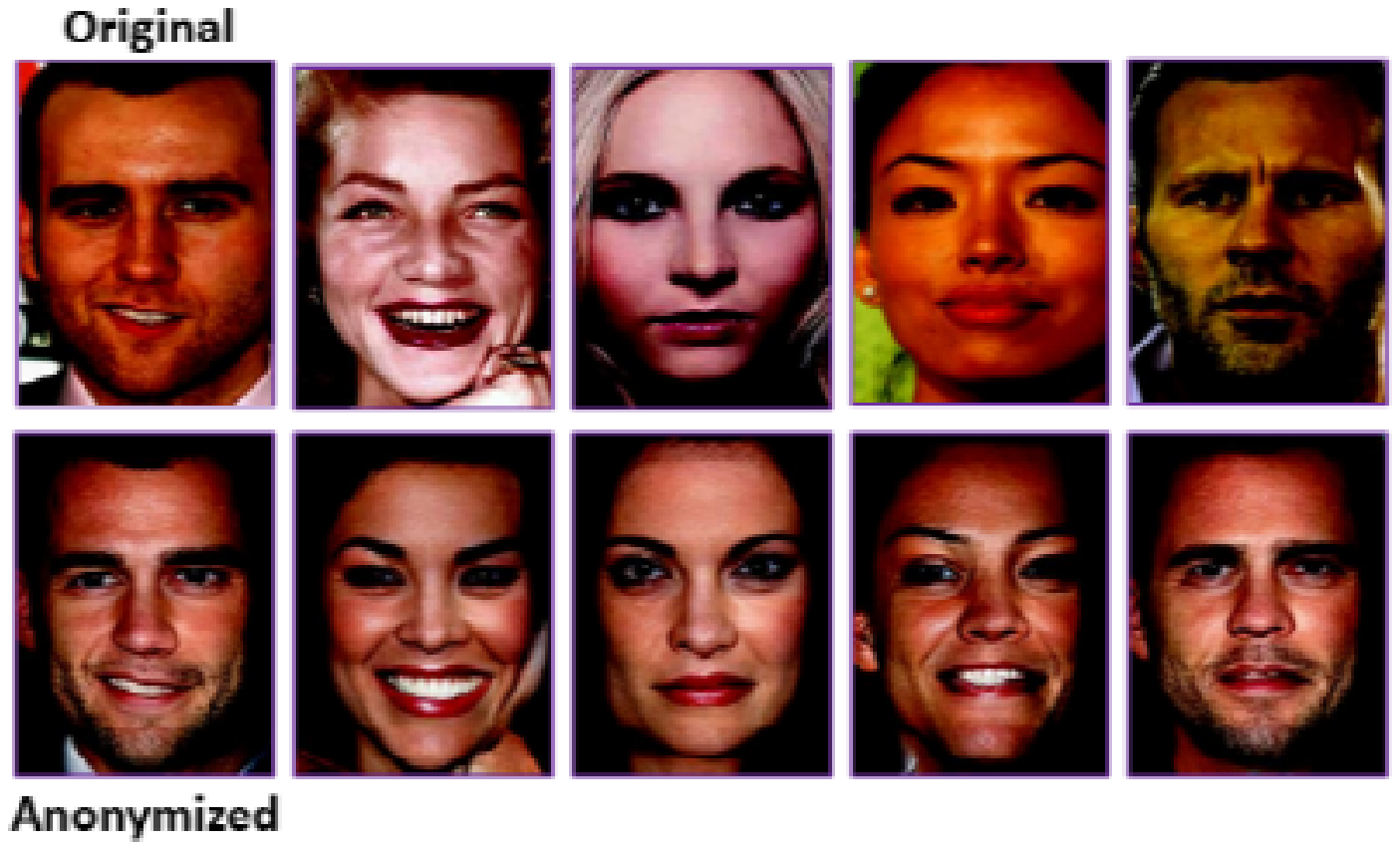}
\vspace{-0.15in}
\caption{Test Results on CelebA-HQ Dataset}
\label{Fig:Result_CelebA-HQ_part}
\vspace{-0.1in}
\end{figure}
\begin{figure}[t]
\centering
\includegraphics[width=0.95\columnwidth]{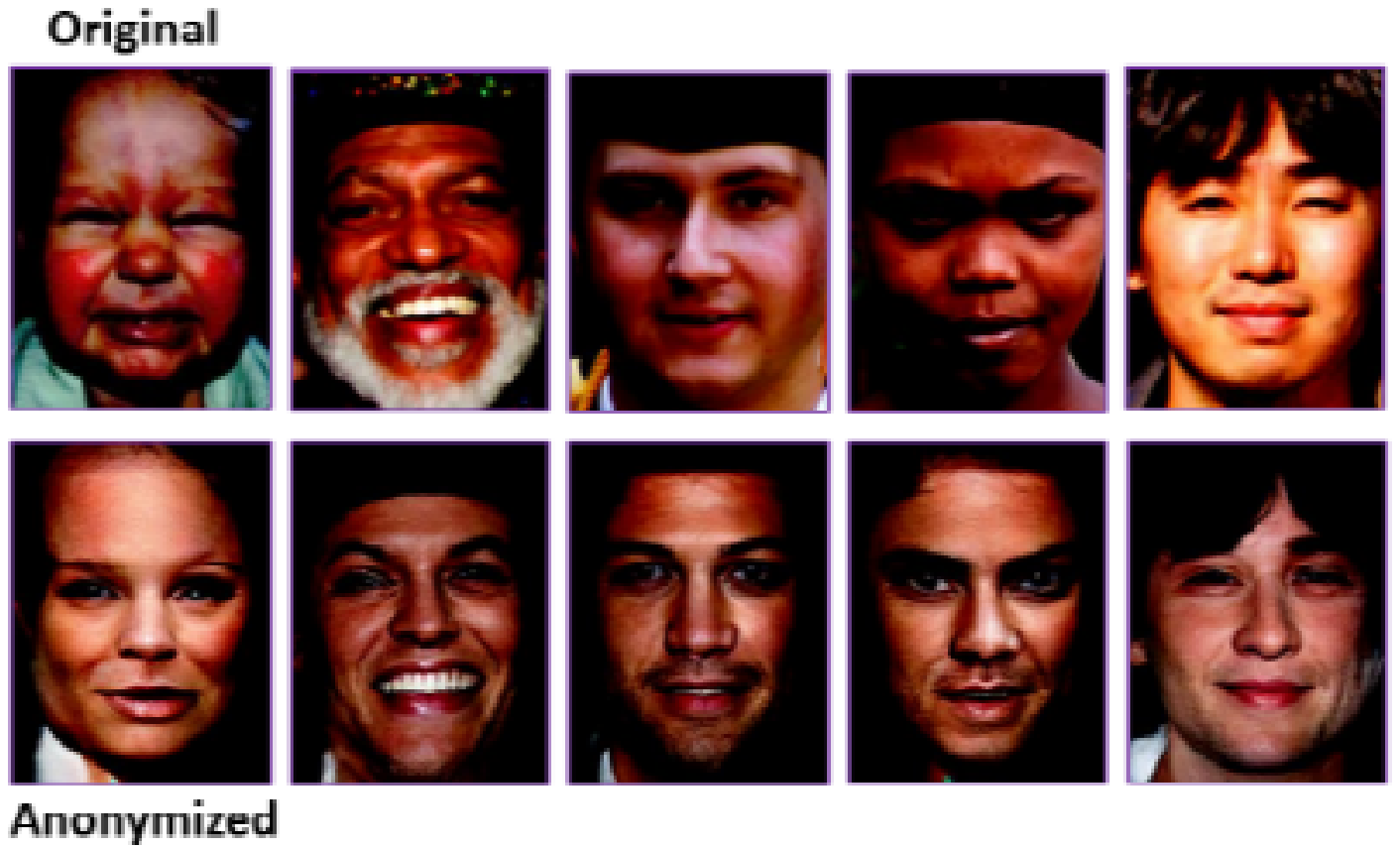}
\vspace{-0.15in}
\caption{Test Results on Helen Dataset}
\label{Fig:Result_Helen_part}
\vspace{-0.1in}
\end{figure}
\begin{figure}[t]
\centering
\includegraphics[width=0.95\columnwidth]{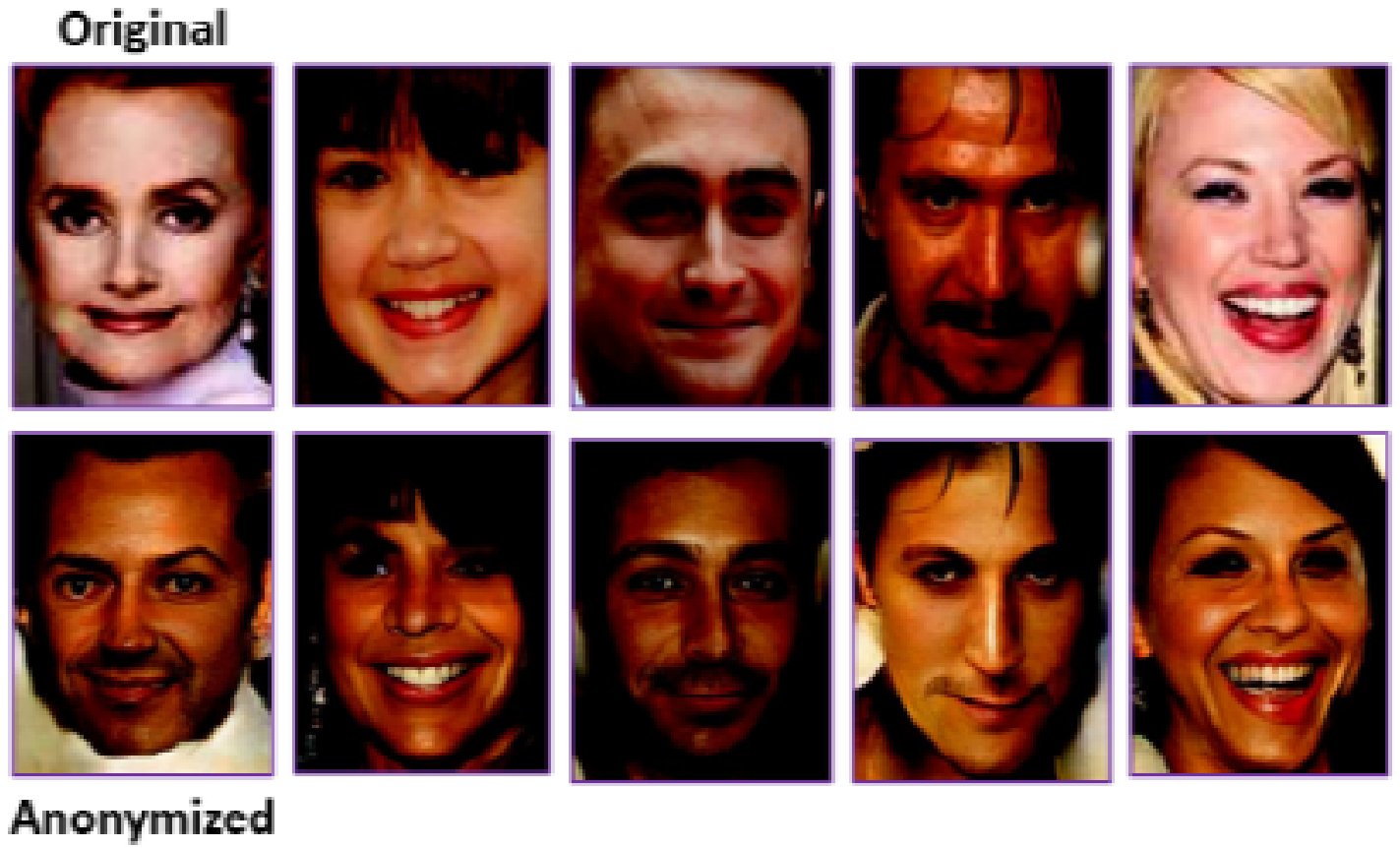}
\vspace{-0.15in}
\caption{Test Results on FaceScrub Dataset}
\label{Fig:Result_FaceScrub_part}
\end{figure}
Figs.~\ref{Fig:Result_CelebA-HQ_part}, \ref{Fig:Result_Helen_part}, and \ref{Fig:Result_FaceScrub_part} provide the quality of our face-anonymizing generators.
We can confirm that our method produces well anonymized faces for diverse faces in CelebA-HQ, Helen, and FaceScrub dataset.
For each dataset, the additional results are shown in Figs.~\ref{Fig:Result_CelebA-HQ}, \ref{Fig:Result_Helen}, and \ref{Fig:Result_FaceScrub}.

\textbf{Quantitative Evaluation of Face Anonymization}:
To evaluate our anonymization system quantitatively, we utilize siamese network \cite{siamese_net}.
The siamese network is widely utilized to measure the dissimilarity between two images.
Via the siamese network, we measure how dissimilar an anonymized face and an original face are. 

For our evaluation, we train a siamese network to calculate Euclidean distance between two faces.
The larger the euclidean distance is, the more dissimilar two faces are.
The inception resnet \cite{InceptionResnet} is adopted for the backbone network of the siamese network.
We perform 50 epochs of training on the CASIA-WebFace dataset \cite{CASIA}, which includes about 500,000 images, and the image sizes are $256 \times 256$.

\begin{table}[t]
\centering
\caption{Average Euclidean distance for each test dataset}
\label{table:siamese_result}
\begin{tabular}{|c||c|c||c|l|c|c|}
\hline
  & \multicolumn{2}{c||}{Criterion} & \multicolumn{4}{c|}{Test Dataset}               \\ \hline
\multirow{3}{*}{\begin{tabular}[c]{@{}c@{}}Average\\ Euclidean\\ Distance\end{tabular}} &
  Same &
  Different &
  \multicolumn{2}{c|}{\multirow{2}{*}{\begin{tabular}[c]{@{}c@{}}CelebA\\ HQ\end{tabular}}} &
  \multirow{2}{*}{Helen} &
  \multirow{2}{*}{FaceScrub} \\
 & person         & people        & \multicolumn{2}{c|}{}     &          &          \\ \cline{2-7} 
 & 1.15           & 2.13          & \multicolumn{2}{c|}{1.57} & 2.59     & 1.94     \\ \hline
\end{tabular}
\end{table}

Table~\ref{table:siamese_result} presents the average Euclidean distance for each test dataset.
In the criterion, the value for 'Same person' is average Euclidean distance between faces of the same person, and another value for 'Different people' is obtained between faces of different people.
Those values are measured during training the siamese network.
For each test dataset, the value is the average Euclidean distance between an original face and an anonymized face by our system.
Note that the larger the value is, the more dissimilar two faces are.

According to Table~\ref{table:siamese_result}, our anonymization system indeed can make a face that looks different from an original face.
For all test dataset, the average Euclidean distance is larger than that of criterion.

\begin{figure}[t]
\centering
\includegraphics[width=1\columnwidth]{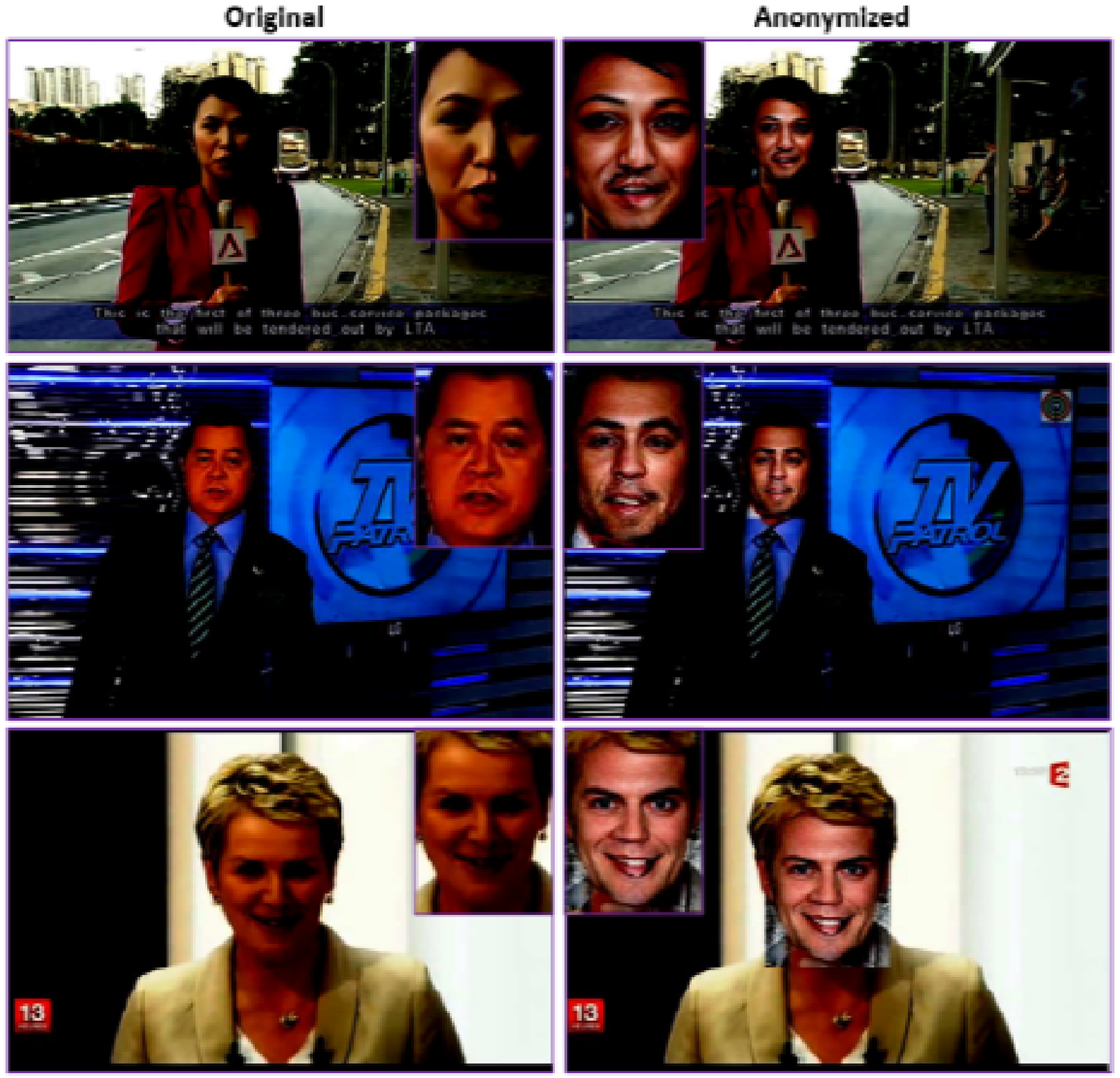}
\caption{Test Results on FaceForensic Dataset via Algorithm~\ref{alg:face_anonymizing_algorithm}}
\label{Fig:Result_FaceForensic_3}
\end{figure}
%
\begin{figure*}[t]
\centering
\includegraphics[width=1.8\columnwidth]{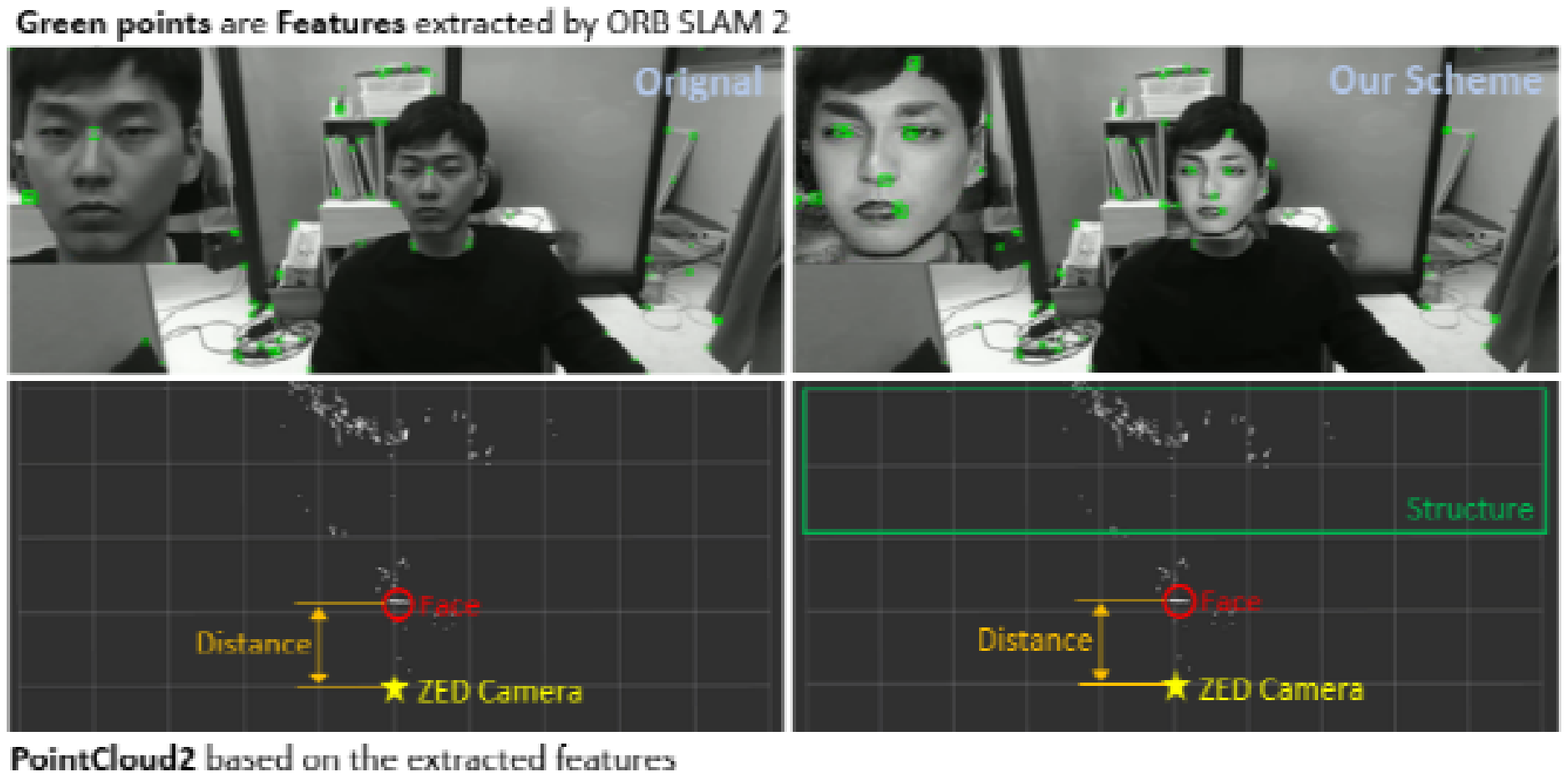}
\caption{SLAM Results on a original video and an anonymized video}
\label{Fig:SLAM_Results_Desk}
\vspace{-0.1in}
\end{figure*}
\begin{figure*}[t]
\centering
\includegraphics[width=2\columnwidth]{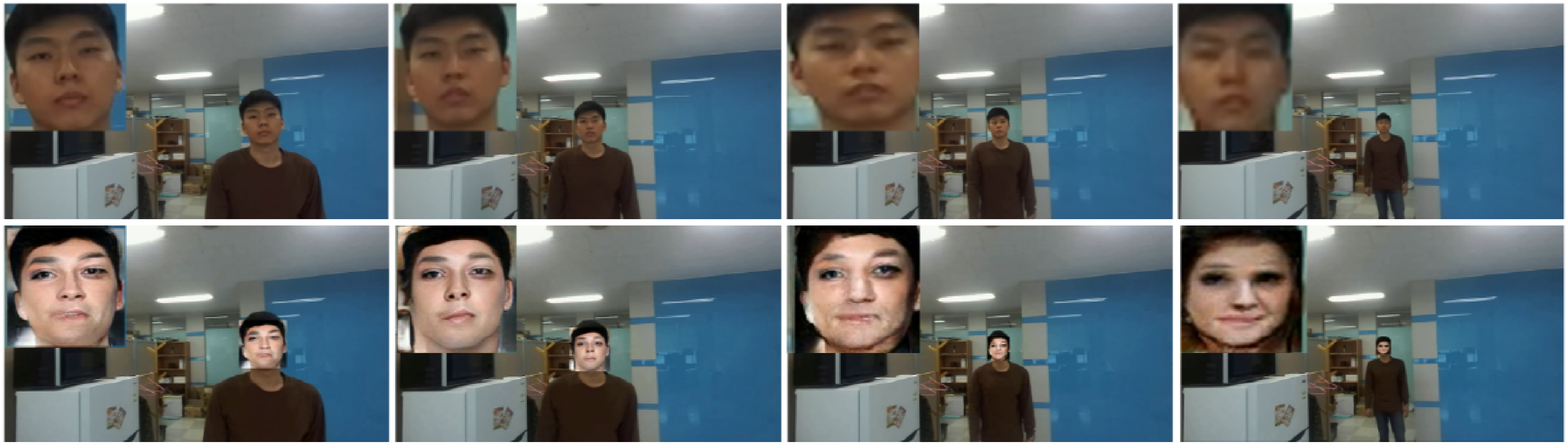}
\caption{Our face-anonymizing system results on drone}
\label{Fig:Evaluation_on_drone}
\vspace{-0.1in}
\end{figure*}
\begin{figure*}[t]
\centering
\includegraphics[width=2\columnwidth]{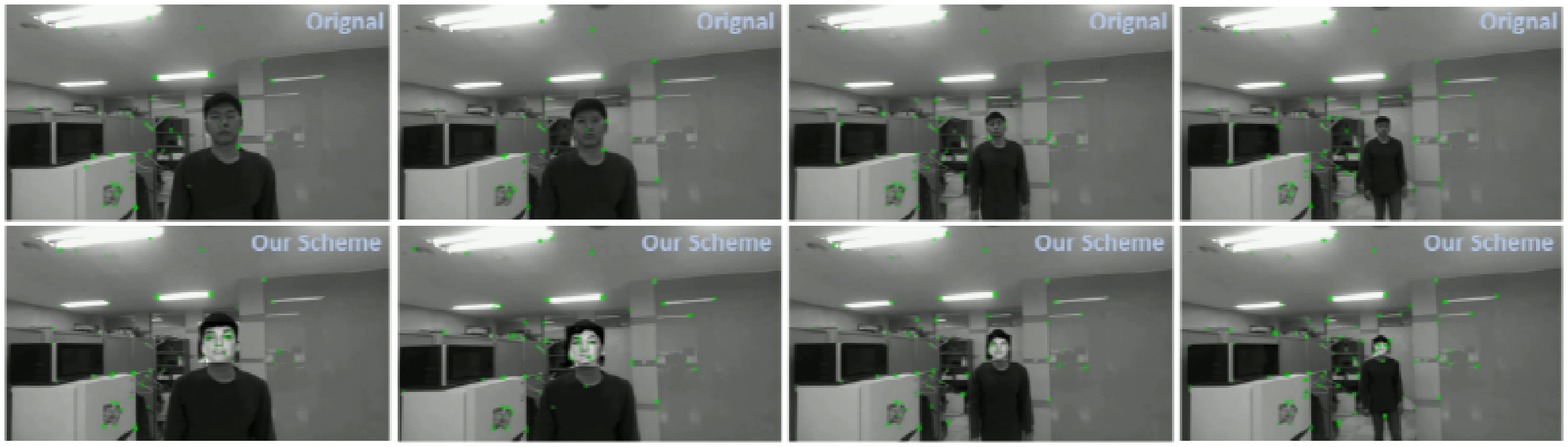}
\caption{SLAM results on drone}
\label{Fig:Evaluation_SLAM_on_drone}
\vspace{-0.1in}
\end{figure*}
\subsection{Evaluation of Algorithm~\ref{alg:face_anonymizing_algorithm}}
\begin{figure*}[t]
\centering
\includegraphics[width=0.7\paperwidth]{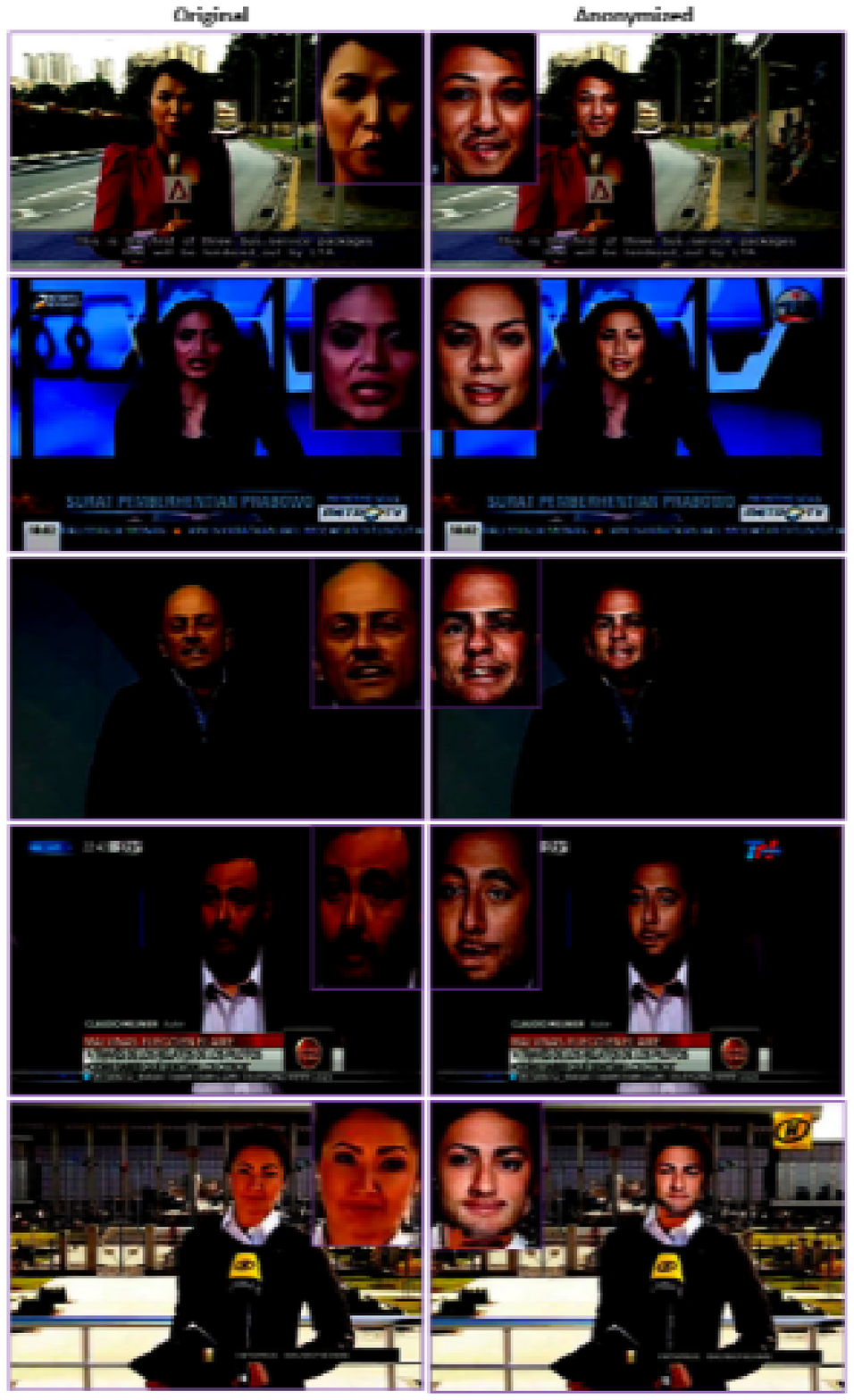}
\caption{Additional Test Results on FaceForensic Dataset via Algorithm~\ref{alg:face_anonymizing_algorithm}}
\label{Fig:Result_FaceForensic_1}
\end{figure*}

\begin{figure*}[t]
\centering
\includegraphics[width=0.7\paperwidth]{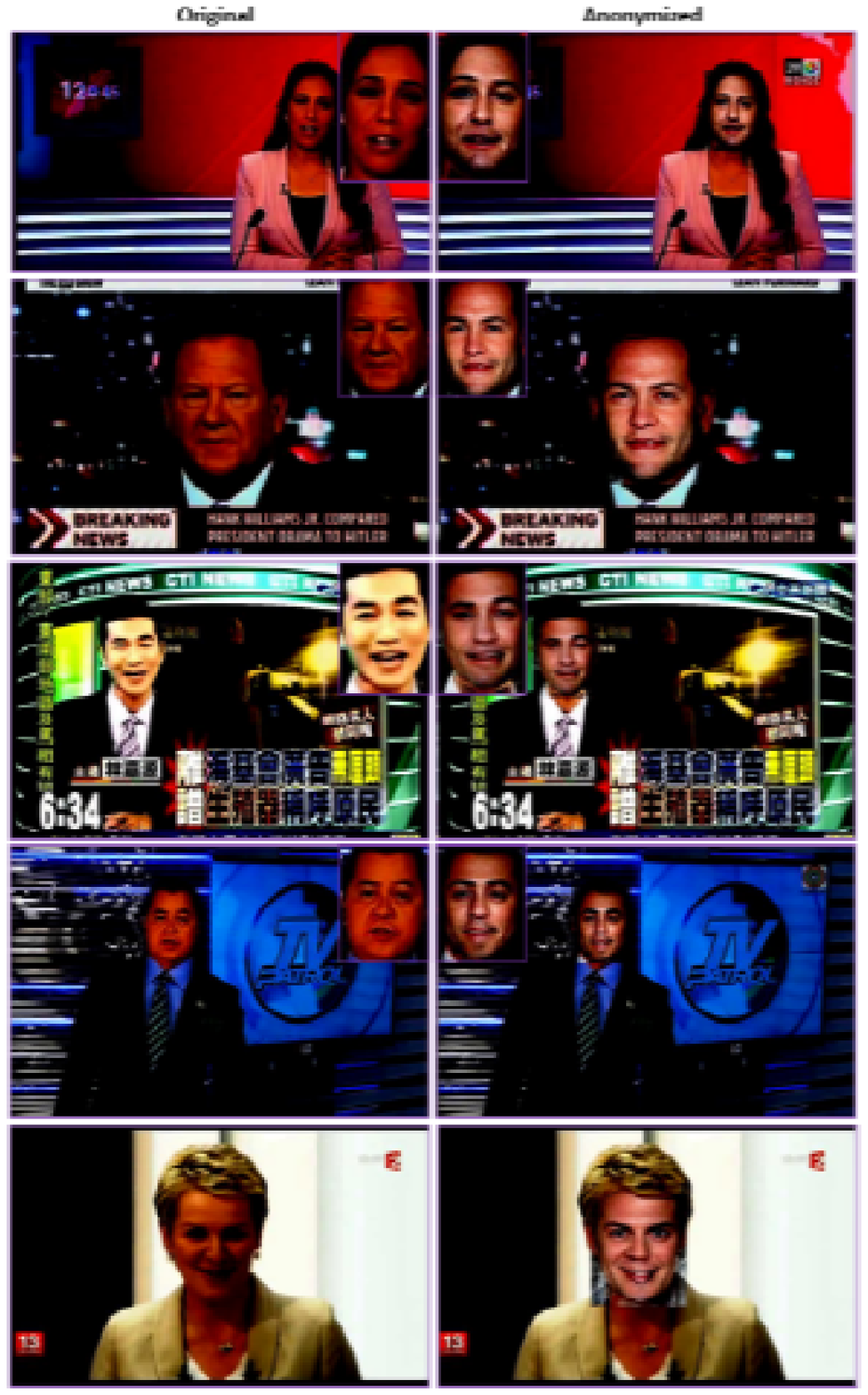}
\caption{Additional Test Results on FaceForensic Dataset via Algorithm~\ref{alg:face_anonymizing_algorithm}}
\label{Fig:Result_FaceForensic_2}
\end{figure*}

To test Algorithm~\ref{alg:face_anonymizing_algorithm}, we conduct our algorithm on FaceForensic video dataset.
Fig.~\ref{Fig:Result_FaceForensic_3} presents results for various videos.
In each result for 'original' and 'Anonymized', a enlarged face image is attached for readability.
Those results confirm that our algorithm finds a face in a video frame, and then anonymizes the detected face well.
The additional results are shown in Figs.~\ref{Fig:Result_FaceForensic_1} and \ref{Fig:Result_FaceForensic_2}.

As a result, for various dataset, our privacy-protection drone vision system can remove privacy information of a person by anonymizing his/her face.

\subsection{SLAM Results}
In this section, we investigate the impact of our proposal on vision-based robot perception, ORB-SLAM2.
Fig.~\ref{Fig:SLAM_Results_Desk} shows two types of results:
\begin{itemize}
    \item \textit{Frames with SLAM's features} are obtained via ORB-SLAM2, in the upper figures of Fig.~\ref{Fig:SLAM_Results_Desk}. The resultant images include green boxes that are feature points extracted by ORB-SLAM2.
    The feature points indicate edges, lines, and corners of objects in a image.
    \item \textit{PointCloud2's results} are drawn via RVIZ \cite{pointcloud2, rviz} based on the extracted feature points.
    In the lower figures of Fig.~\ref{Fig:SLAM_Results_Desk}, the white points are stamped by the calculated distance between a ZED camera and extracted feature points.
\end{itemize}

In Fig.~\ref{Fig:SLAM_Results_Desk}, the upper figures present the comparison of feature extraction results on a original video frame and an anonymized video frame.
In both frames, the feature points are well extracted.
The extracted feature points of our anonymization scheme are almost the same as that of the original video.
In addition, we can notice that the face is also anonymized well by our scheme.

The lower figures present the distance between a camera and a face, measured by ORB-SLAM2.
In the red circle, the white points are extracted from the face.
Hence, the distance between those points and a ZED camera means the distance between the face and the camera.
Under our scheme, the distance is almost the same as that under the original video.

In summary, our anonymization scheme has no impact on the performance of ORB-SLAM2, which confirms that our system can preserve vision-based drone perception well.
In the drone perception, inaccuracy can indeed incur an accident where a drone hits a person's face.
Since our approach effectively protects the person's privacy with good perception, our anonymization scheme is certainly proper to privacy-protection drone patrol system.

\subsection{Drone Hardware Setup}
\textbf{Drone Hardware Specification}: 
We build a customized drone each component of which is as follows.
We selected DJI F550 for our drone frame.
1137 T-motor V2 carbon fiber propeller and T-motor MN3110 KV 780 are selected as propellers and motors.
T-motor Air 40A is chosen as electronic speed controllers.
We utilize Holybro Pixhawk4 for our drone control computer.
Finally, we selected ZED stereo camera for our high-resolution camera.

\textbf{Experiments of Face-Anonymizing Networks with Nvidia Xavier}: 
NVIDIA Jetson Xavier is an machine-learning computer for autonomous machines with the high-computing power.
Especially, this companion computer has a 512-core GPU, and thus it is indeed suitable for large-scale matrix operations which are needed for efficient neural networks computation.
A 8800mah 4S1P LiPo battery was installed for power supply, which provides a voltage of 14.8V.

\subsection{Our Face-Anonymizing System Result on Drone}
Fig.~\ref{Fig:Evaluation_on_drone} shows our face-anonymizing system results on a video recorded in a drone.
The drone is hovering in our laboratory, and a person is walking in front of the drone.
From this result, we can confirm that our system can well anonymize a face with varying its size.

Fig.~\ref{Fig:Evaluation_SLAM_on_drone} presents the comparison of feature extraction results on an original video and an anonymized video.
Both videos confirm that the feature points are well extracted.
In our anonymization scheme, the extracted feature points are almost the same as that of the original video.


%
%

\section{Conclusion}
\label{sec:Conclusion}

In this paper, we developed a privacy-protection drone patrol system with face-anonymizing deep-learning networks.
To train face-anonymizing networks, we proposed a training architecture, which consists of the segmentation learning part and the synthesis learning part.
The training architecture is constructed by combining CycleGAN and GauGAN. 
We additionally modified generators' losses of those GANs for our purpose.
In our system, our face-anonymizing networks transform all original faces in every snapshot to different faces.
Hence, in snapshots, privacy of people can be fundamentally protected.
Via various test dataset, we confirmed that our system can indeed preserve the person's privacy qualitatively and quantitatively.
In addition, by implementing ORB-SLAM2, we also verified that our system can preserve the vision-based perception of drone with well-anonymized faces.
Finally, our system was also evaluated with actually recorded videos on drone.


\bibliographystyle{ieeetr}
\bibliography{06_references}

\begin{thebibliography}{10}
\bibitem{Privacy_Infringement_1}
J.~Littman, ``7 sightings that prove the robot invasion is already here.''
\end{thebibliography}

\end{document}